\documentclass[journal]{IEEEtran}
\usepackage{array}
\usepackage{textcomp}
\usepackage{stfloats}
\usepackage{url}
\usepackage{verbatim}
\usepackage{cite}
\usepackage{wrapfig}
\usepackage{subcaption}
\usepackage{graphicx}
\usepackage{mathptmx}
\usepackage{booktabs}
\usepackage{algorithm}
\usepackage{algpseudocode}
\usepackage{latexsym}
\usepackage{amsmath,amssymb}
\usepackage{mathtools}
\usepackage{stmaryrd}
\usepackage{bm}
\usepackage{bbm}
\usepackage{amsthm}
 
\providecommand{\argmin}{\operatornamewithlimits{argmin}}

\DeclareMathOperator{\Var}{Var}

\providecommand{\R}{\mathbb{R}} 
\providecommand{\E}{\mathbb{E}} 
\providecommand{\T}{\mathrm{T}} 

\providecommand{\ind}[1]{\ensuremath{\mathbbm{1}{\left\{#1\right\}}}} 
\renewcommand{\geq}{\geqslant}
\renewcommand{\leq}{\leqslant}
\DeclarePairedDelimiterX{\inner}[2]{\langle}{\rangle}{#1, #2}
\DeclarePairedDelimiter{\norm}{\lVert}{\rVert}
\DeclarePairedDelimiter{\abs}{\lvert}{\rvert}
\DeclarePairedDelimiter{\rbra}{(}{)}
\DeclarePairedDelimiter{\sbra}{[}{]}
\DeclarePairedDelimiter{\cbra}{\{}{\}}
\usepackage[colorlinks=true,linkcolor=blue,pdfborder={0 0 0}]{hyperref}
\usepackage{cleveref}
\newtheorem{theorem}{Theorem}[]
\newtheorem{proposition}[theorem]{Proposition}
\newtheorem{corollary}[theorem]{Corollary}
\newtheorem{lemma}[theorem]{Lemma}
\theoremstyle{definition}
\newtheorem{definition}[]{Definition}

\newtheorem{assumption}[]{Assumption}
\usepackage{ifthen}

\usepackage{multirow}
\newcommand{\markupdraft}[2]{
\ifthenelse{\equal{#1}{display}}{#2}{}
\ifthenelse{\equal{#1}{color}}{\color{#2}}{}
}

\newcommand{\newcolored}[3][]{{\markupdraft{color}{#2}#3}
\ifthenelse{\equal{#1}{}}{}{\markupdraft{display}{{\color{yellow!70!black}[#1]}}}}
\newcommand{\del}[2][]{{\markupdraft{display}{{\color{orange}[removed: ``#2''[#1]]}}}} 
\newcommand{\new}[2][]{\newcolored[#1]{blue}{#2}}

\renewcommand{\del}[2]{}  
\renewcommand{\markupdraft}[2]{}  

\newcommand{\iu}{\mathrm{i}\mkern1mu}  

\begin{document}

\title{
Theoretical Analysis of Explicit Averaging and Novel Sign Averaging in Comparison-Based Search 
\thanks{This study was partially supported by JSPS KAKENHI under Grant Number 19H04179.}
}
\author{
Daiki Morinaga,
\thanks{D. Morinaga is with the Department of Computer Science, University of Tsukuba, Tsukuba, Japan; RIKEN AIP, Japan (e-mail: morinaga@bbo.cs.tsukuba.ac.jp).}
\and Youhei Akimoto\thanks{
Y. Akimoto is with the Institute of Systems and Information Engineering, University of Tsukuba, Tsukuba, Japan; RIKEN AIP, Japan
(e-mail: akimoto@cs.tsukuba.ac.jp).}
}

\markboth{Journal of \LaTeX\ Class Files, }%
{Shell \MakeLowercase{\textit{et al.}}: A Sample Article Using IEEEtran.cls for IEEE Journals}

\maketitle
\begin{abstract}
In black-box optimization, noise in the objective function is inevitable.
Noise disrupts the ranking of candidate solutions in comparison-based optimization, possibly deteriorating the search performance compared with a noiseless scenario.
Explicit averaging takes the sample average of noisy objective function values and is widely used as a simple and versatile noise-handling technique. 
Although it is suitable for various applications, it is ineffective if the mean is not finite.
We theoretically reveal that explicit averaging has a negative effect on the estimation of ground-truth rankings when assuming stably distributed noise without a finite mean. 
Alternatively, sign averaging is proposed as a simple but robust noise-handling technique.
We theoretically prove that the sign averaging estimates the order of the medians of the noisy objective function values of a pair of points with arbitrarily high  probability as the number of samples increases.
Its advantages over explicit averaging and its robustness are also confirmed through numerical experiments.
\end{abstract}


\sloppy

\section{Introduction}

    Evolutionary computation approaches in the continuous domain are intended to approximate the optimal solution that minimizes objective function $f: \R^D \to \R$ through queries $x \mapsto f(x)$ computed by (possibly expensive) numerical simulations for each design variable $x$.
    With the rapid development of computational performance, such approaches have been increasingly exploited in practical and industrial applications, such as geoscience~\cite{kriest2017calibrating}, topology optimization~\cite{fujii2018cma}, and machine learning~\cite{dong2019efficient}.
    A major characteristic of most evolutionary computation approaches, such as the evolution strategy with covariance matrix adaptation (CMA-ES)~\cite{hansen2006cma}, and some deterministic approaches, such as the Nelder--Mead method~\cite{nelder1965simplex}, is their comparison-based principle. 
    Instead of relying on the objective function values, these approaches use the order of these values from the population of design variables. 
    Although the direct use of the objective function value has been widely studied (e.g., in stochastic zeroth-order optimization~\cite{ghadimi2013stochastic}),
    comparison-based approaches provide invariance against the strictly increasing transformations of the objective function~\cite{auger2016linear}.
    Under invariance, any convergence statement for a single function can be generalized, both theoretically and experimentally, to a class of functions that are transformations of the original function.
    Furthermore, while the non-differentiability or discontinuity of the objective function is often troublesome for value-based approaches, it may be avoided if a strictly increasing transformation exists for converting a troublesome problem into a feasible one.

    Real-world applications often entail the presence of noise in the $f$ values.
    Consequently, each query $x \mapsto f(x; \bm{\epsilon})$ returns the objective function value corrupted by noise $\bm{\epsilon} \in \mathrm{R}^M$, which is independently sampled per query but remains concealed. 
    Thus, the optimization objective must consider the noise distribution, which is often defined as
    \begin{equation}
        \argmin_{x\in\R^D} \mathcal{A}[f(x; {\bm \epsilon})]
        ,
    \end{equation}
    where $\mathcal{A}$ denotes an optimization index. 
    The choice of $\mathcal{A}$ depends on measures such as the expectation, percentile, or conditional value at risk~\cite{gabrel2014recent, rakshit2017noisy}.
    In comparison-based search, the uncertainty of the objective function affects the population rankings. 
    Therefore, given a population of design variables, $\{x_i\}_{i = 1}^\lambda$, the search algorithms must determine the population rankings with respect to the ground-truth objective value, $\{\mathcal{A}[f(x_i; {\bm \epsilon})]\}_{i=1}^{\lambda}$. 
    However, because search algorithms observe noisy objective function values, ground-truth ranking cannot be achieved, thereby deteriorating the algorithm performance.

    Various noise-handling techniques have been proposed to address uncertainty in evolutionary computation. These techniques can be classified into five groups~\cite{rakshit2017noisy}: 1) estimating the ground truth by resampling and averaging the queries (i.e., explicit averaging), 2) estimating the ground truth without using explicit averaging, 3) enlarging the population size (implicit averaging), 4) improving the search strategy, and 5) modifying the selection strategy.
    Although no existing technique is predominant~\cite{arnold2001local, jin2005evolutionary, arnold2006general, ahrari2022revisiting, qian2018effectiveness, beyer2007robust}, explicit averaging is often employed in practice because it can be seamlessly incorporated into most evolutionary computation approaches.
    Explicit averaging resamples the noisy objective function values over $K \geq 1$ times per $x_i$ and determines the population rankings, $\{x_i\}_{i=1}^\lambda$, based on the average objective function values, $\big\{\frac{1}{K} \sum_{k=1}^{K} f(x_i; \bm{\epsilon_{i,k}}) \big\}$. 
    Considering the strong law of large numbers, the averaged values converge to $\{\E[f(x_i; {\bm \epsilon})]\}_{i=1}^{\lambda}$ almost surely if the distribution of the noisy objective function values has a finite mean. Therefore, explicit averaging seems reasonable for handling noise in the expectation as the optimization index.\footnote{
        Explicit averaging has a main advantage over implicit averaging. 
        Assume that the ground-truth value satisfies $f_\mathrm{GT}(x) = \E[f(x; {\bm \epsilon})] < \infty$.
        Implicit averaging attempts to estimate the ranking of the objective function values, $\{f(x_i; {\bm \epsilon})\}_i^\lambda$, in population $\{x_i\}_i^\lambda$.
        The algorithm estimation of the population ranking can be evaluated using the Kendall rank correlation coefficient, $\tau$, which measures the rank correlation between two sets of sample values  $\{f(x_i; {\bm \epsilon}_i)\}_i^\lambda$ and ground truth $\{\E[f(x_i; {\bm \epsilon})]\}_i^\lambda$.
        Coefficient $\tau$ is defined as
        \begin{equation*}
            \tau := \frac{2}{\binom{\lambda}{2}} \cdot \sum_{i>j}^\lambda\ind{(f(x_i; {\bm \epsilon}_i)-f(x_j;{\bm \epsilon}_j))
            (\E[f(x_i; {\bm \epsilon})] - \E[f(x_j; {\bm \epsilon})])>0} - 1
            .
        \end{equation*}
        Let $p := \Pr_{x, {\bm \epsilon}}\sbra*{(f(x_i; {\bm \epsilon}_i)-f(x_j;{\bm \epsilon}_j)(\E[f(x_i; {\bm \epsilon})] - \E[f(x_j; {\bm \epsilon})])>0}<1$. 
        Considering the law of large numbers, $\lim_{\lambda\to\infty} \tau = 2p  - 1<1$.
        Therefore, implicit averaging cannot provide an exact estimate of ranking $\tau=1$, even in the limit $\lambda\to\infty$. In contrast, with explicit averaging, $\frac{1}{K}\sum_k^K f(x; {\bm \epsilon}_k)\to\E[f(x; {\bm \epsilon})]$ as $K\to\infty$ given the law of large numbers.}

    We theoretically investigate the effect of explicit averaging on comparison-based algorithms.
    Existing theoretical studies on explicit averaging in continuous black-box optimization often assume a finite variance for the noisy objective function~\cite{beyer1993toward, beyer2000evolutionary,beyer2006functions, hansen2008method, qian2018effectiveness}.
    For finite variance, explicit averaging reduces the standard deviation of the distribution of the noisy objective function values by a factor of $1/\sqrt{K}$. 
    However, it remains unclear whether explicit averaging is effective for heavy-tailed distributions of objective function values. 
    In discrete optimization, Doerr and Sutton~\cite{doerr2019resampling} showed that averaging may fail when the noise follows a Cauchy distribution, which does not have finite mean and variance.
    From these observations and the strong law of large numbers holding even for infinite variance, we conjecture that explicit averaging is effective if the noise distribution has a finite mean, whereas it does not work otherwise. 
    To the best of our knowledge, no theoretical study has been conducted on the impact of explicit averaging on a comparison-based search algorithm in the continuous domain for a heavy-tailed noise distribution.
    
    We consider two main research questions. 1) How effectively does averaging with sample size $K$ work on comparison-based optimization algorithms when the noise distribution does not have the $p$-th absolute moment as a finite value? 2) Can we design a noise-handling technique with theoretical grounds for a noise distribution without a finite mean?

    The main contributions of this study are summarized as follows:
    \begin{itemize}
    \item 
    In \Cref{sec:oep}, the effectiveness of explicit averaging with the stable noise is addressed.
    Under \Cref{asm:function} (linearity of $f(x; {\bm \epsilon})$ over ${\bm \epsilon}$) and \Cref{asm:noise} (stable noise), \Cref{theo:oep_stabledist} reveals the range of stable parameter $\alpha$ that determines whether explicit averaging works. For $\alpha\in(1, 2]$, explicit averaging is effective, while for $\alpha=1$, it has no effect, and for $\alpha\in(0, 1)$, explicit averaging is even harmful.
    For stable noise, $\alpha$ determines the finiteness of the $p$-th absolute moment.
    \item
    In \Cref{sec:sign_averaging}, we propose a variant of explicit averaging called \emph{sign averaging}.
    In \Cref{theo:oep_stabledist_sign}, we show that sign averaging is always effective under different assumptions, namely, \Cref{asm:symmetric} (additivity of the median) and \Cref{asm:uniqueness} (continuity of the distribution of $f(x; {\bm \epsilon})$).
    If the distribution of $f(x; {\bm \epsilon})$ is symmetric, these two assumptions are satisfied under \Cref{asm:function} and \Cref{asm:noise}.
    \end{itemize}

\section{Averaging Analysis}\label{sec:oep}

        We theoretically analyze the effect of increasing the sample size, $K$, by averaging $\frac{1}{K}\sum_k^K f(x; {\bm \epsilon}_k)$ on the \emph{order estimation probability} (OEP), which is the probability of the estimate of the order of two solutions, $x_1$ and $x_2$, agreeing with the ground truth.

\subsection{Stable Distribution}\label{sec:stabledist}

        In this section, we assume that the noise follows a \emph{stable distribution}, as we formally state below. 
        The stable distribution is considered a generalization of the well-studied distribution in noisy optimization, namely, the \emph{normal distribution}, because the stable distribution is the only possible distribution of the limit of the sum of a sequence of independent and identically distributed (i.i.d.)\ random variables considering the generalized central limit theorem \cite{nolan2020stable}. 

        Stable distribution ${\bm S}(\alpha, \beta, \gamma, \delta)$ is formally defined as follows.
        \begin{definition}[Stable distribution ${\bm S}(\alpha, \beta, \gamma, \delta)$ in Definition 1.8 of \cite{nolan2020stable}]\label{def:stable_dist1}
            Let $\alpha\in(0, 2]$, $\beta\in[-1, 1]$, $\gamma\in(0, \infty)$, and $\delta\in\R$.
            In addition, let random variable $X$ with support $\R$ have the following characteristic function:
            \begin{multline}
                \E[\exp(itX)] = 
                \\
                    \begin{cases}
                    \exp\left(-\gamma^\alpha|t|^\alpha \left(1-\iu\beta\tan\left(\frac{\pi\alpha}{2}\right)\mathrm{sign}(t)\right)+\iu\delta t\right)
                    & \text{ for } \alpha\neq 1; \\
                    \exp\left(-\gamma|t| \left(1+\iu\beta\frac{2}{\pi}\mathrm{sign}(t)\log|t|\right)+\iu\delta t\right)
                    & \text{ for } \alpha=1.
                    \end{cases}
                \label{eq:charfunc_stable}
            \end{multline}
            The probability distribution of $X$ is called a stable distribution and denoted by ${\bm S}(\alpha, \beta, \gamma, \delta)$, where $\iu$ is the imaginary unit.
        \end{definition}
        Parameters $\alpha$, $\beta$, $\gamma$, and $\delta$ are often referred to as stability parameter, skewness parameter, scale parameter, and location parameter, respectively.

        Stable distributions include well-known distributions. 
        For example, ${\bf S}(\alpha=2, \beta, \gamma, \delta)$ is a normal distribution with mean $\delta$ and variance $2\gamma^2$, 
        while ${\bf S}(\alpha=1, \beta=0, \gamma, \delta)$ is a Cauchy distribution with mode $\delta$,
        and the Levy distribution is obtained as ${\bf S}(\alpha=1/2, \beta=1, \gamma, \delta)$.
        The skewness parameter controls the symmetricity of the distribution. If $\beta=0$, the distribution is symmetric.
        Note that if $\alpha=2$, $\beta$ is ignored and ${\bf S}(2, \beta, \gamma, \delta)$ is always symmetric.
        
        An important property of this distribution is integrability.
        The variance exists as a finite value if and only if $\alpha = 2$, that is, a normal distribution.
        For $0<\alpha<2$, the $p$-th absolute moment, $\E[|X|^p]$, is finite only for $0<p<\alpha$~\cite{nolan2020stable}.
        Therefore, not even the mean is defined for $0<\alpha\leq 1$.

        The stable distribution is said to be \emph{stable} because a linear combination of two random variables following stable distributions with the same $\alpha$ is again a random variable following a stable distribution with the same $\alpha$. This property is formally stated as follows.
        \begin{proposition}[Linear transformation of stable distribution in Proposition 1.17 of \cite{nolan2020stable}]
        \label{prop:stabledist_transformation1}
        The following properties hold for stable distributions:
        \begin{enumerate}
            \item 
            If $X\sim{\bm S}(\alpha, \beta, \gamma, \delta)$, then for any $a, b\in\R$, 
            \begin{multline}
            aX+b\sim
            \\
            \begin{cases}
                {\bm S}(\alpha, \mathrm{sign}(a)\cdot\beta, |a|\gamma, a\delta+b)
                &
                \alpha\neq 1
                ;
                \\
                {\bm S}\left(1, \mathrm{sign}(a)\cdot\beta, |a|\gamma, a\delta+b - \frac{2}{\pi}\beta\gamma a\log|a|\right)
                &\alpha=1
                .
            \end{cases}
            \end{multline}
            \item
            If $X_1\sim{\bm S}(\alpha, \beta_1, \gamma_1, \delta_1)$ and 
             $X_2\sim{\bm S}(\alpha, \beta_2, \gamma_2, \delta_2)$ are independent, then $X_1+X_2\sim{\bm S}(\alpha, \beta, \gamma, \delta)$, where
             \begin{equation}
                 \beta = \frac{\beta_1\gamma_1^\alpha + \beta_2\gamma_2^\alpha}{\gamma_1^\alpha+\gamma_2^\alpha}
                 ,\enspace
                 \gamma = \rbra*{\gamma_1^\alpha + \gamma_2^\alpha}^{\frac{1}{\alpha}}
                 ,\enspace
                 \delta = \delta_1 + \delta_2
                 .
             \end{equation}
        \end{enumerate}
        \end{proposition}

    \subsection{Problem Statement}\label{subsec:assumption}

    We assume that $f(x; {\bm \epsilon})$ is linear with respect to ${\bm \epsilon}$ and that each element of ${\bm \epsilon} = [\epsilon_1, \dots, \epsilon_M]^\T$ follows a stable distribution, as formally stated below. 
        \begin{assumption}\label{asm:function}
            Function $f: \R^D \times \R^M \to \R$ is defined as
            \begin{equation}
                f(x; {\bm \epsilon}) =
                h(x) + \sum_m^M g_m(x)\cdot \epsilon_m
                \label{eq:asm_f}
            \end{equation}
        for some $h:\R^D\to\R$ and $g_m:\R^D\to\R$ with $m = 1, 2, \ldots, M$.
        \end{assumption}
        \begin{assumption}\label{asm:noise}
            Noise vector ${\bm \epsilon} = [\epsilon_1, \dots, \epsilon_M]^\T$ consists of random variables that follow a stable distribution with a common $\alpha \in (0, 2]$. The distribution of the $m$–th element, $\epsilon_m$, of ${\bm \epsilon}$ is ${\bf S}(\alpha, \beta_m, \gamma_m, \delta_m)$.
        \end{assumption}

        The problem classes defined by \Cref{asm:function} and \Cref{asm:noise} include frequently studied problem instances. A function with additive noise $f(x; {\bm \epsilon}) = h(x) + \epsilon_1$ is included when $M = 1$ and $g_1(x) = 1$. A function with multiplicative noise $f(x; {\bm \epsilon}) = h(x) (1 + \epsilon_1)$ is included when $M = 1$ and $g_1(x) = h(x)$. Noise is often assumed to follow a normal distribution, which is an instance of a stable distribution. 
	
        The studied optimization index is given by
        \begin{equation}
            \mathcal{A}[f(x; {\bm \epsilon})]
            = f(x; \Delta)
            ,
        \end{equation}
        where $\Delta = [\delta_1, \dots, \delta_M]^\T$ collects the location parameter, $\delta_m$, of $\epsilon_m$.
       This choice is explained as follows. Arguably, the most common optimization index is expected value $\E_{\bm \epsilon}[f(x; {\bm \epsilon})]$, and explicit averaging likely minimizes the expected value. 
        However, this is undefined for $\alpha \leq 1$ unless $g_m(x) = 0$ for all $m = 1, \dots, M$. 
        If $\alpha \in (1, 2]$, we have $f(x; \Delta) = h(x) + \sum_m^M g_m(x)\cdot \delta_m = \E_{{\bm \epsilon}}[f(x; {\bm \epsilon})]$~\cite{nolan2020stable}.
        Optimization index $f(x; \Delta)$ is well-defined for $\alpha \in (0, 1]$, whereas $\E_{\bm \epsilon}[f(x; {\bm \epsilon})]$ is undefined. 
        Additionally, if $\beta_m = 0$ for all $m = 1, \dots, M$, $f(x; \Delta)$ corresponds to the median of $f(x; \Delta)$ for all $\alpha \in (0, 2]$. 
        Therefore, our objective can be also understood as minimizing the median of the objective function value when $\beta_m = 0$ for all $m = 1, \dots, M$. Hereafter, $\nabla_{\bm \epsilon} f(x; {\bm \epsilon}) = [g_1(x), \ldots, g_M(x)]^\mathrm{T}$ is denoted by $\nabla_{\bm \epsilon} f(x)$ to avoid confusion because it is independent of ${\bm \epsilon}$.

    \subsection{OEP}\label{subsec:oep}

        Comparison-based search algorithms determine the \emph{order} of any pair of candidate solutions $(x_1, x_2)$. 
        For a stochastic objective function, the ground-truth order determined by $f(x_1; \Delta)$ and $f(x_2; \Delta)$ is unknown and needs to be estimated from finite samples $\{f(x_1; {\bm \epsilon}_{1,k})\}_{k=1}^{K}$ and $\{f(x_2; {\bm \epsilon}_{2,k})\}_{k=1}^{K}$, where ${\bm \epsilon}_{i,k}$ are all i.i.d.\ copies of ${\bm \epsilon}$ satisfying \Cref{asm:noise}. 
        A common estimation approach is \emph{explicit averaging}, in which the order of $f(x_1; \Delta)$ and $f(x_2; \Delta)$ is estimated by using the order of $\frac{1}{K} \sum_{k=1}^{K} f(x_1; {\bm \epsilon}_{1,k})$ and $\frac{1}{K} \sum_{k=1}^{K} f(x_2; {\bm \epsilon}_{1,k})$ with $K$ possibly adapted during search.

        We analyze the OEP, which is the probability of correctly estimating the orders of $f(x_1; \Delta)$ and $f(x_2; \Delta)$. The OEP is formally defined as follows.
        Let $\mathrm{sign}(\cdot)$ denote the sign function whose output ($-1, 0$, or $1$) is the sign of its argument.
        \begin{definition}[OEP for $(x_1, \Delta)$ and $(x_2, \Delta)$]\label{def:oep}
            Let $\widehat\eta_{{\bm \epsilon}}(x_1, x_2)\in \{-1, 0, 1\}$ be an estimate of 
            \begin{equation}
            \eta_\Delta(x_1, x_2) := \mathrm{sign}(f(x_1; \Delta) - f(x_2; \Delta))
            .
            \label{eq:eta}
            \end{equation}
            The \emph{OEP} for $(x_1, \Delta)$ and $(x_2, \Delta)$ is given by
            \begin{equation}
                \Pr_{\bm \epsilon}[\widehat\eta_{\bm \epsilon} (x_1, x_2) = \eta_\Delta(x_1, x_2)]
                .
            \end{equation}
        \end{definition}

        The order estimate using explicit averaging is defined as
        \begin{equation}
        \widehat\eta_\epsilon^\mathrm{AVE}(x_1, x_2) := \mathrm{sign}\left(\frac{1}{K}\sum_k^K f(x_1; {\bm \epsilon}_{1, k}) - \frac{1}{K}\sum_k^K f(x_2; {\bm \epsilon}_{2, k})\right)
        .
        \label{eq:eta_ave}
        \end{equation}
        Under \Cref{asm:function} and \Cref{asm:noise}, the argument of $\mathrm{sign}(\cdot)$, namely, $\frac{1}{K}\sum_{k=1}^K f(x_1; {\bm \epsilon}_{1, k}) - \frac{1}{K}\sum_{k=1}^K f(x_2; {\bm \epsilon}_{2, k})$, is stably distributed as stated below.
        Let the diagonal matrix whose $(n, n)$-th element is $X_n$ be denoted by $\mathrm{diag}(X_1, \ldots, X_N)$ and $\norm{X}_\alpha = \rbra{\sum_n^N |X_n|^\alpha}^{1/\alpha}$ for $X = [X_1, \ldots, X_N]^\mathrm{T}$ be the $\alpha$-norm for $\alpha \geq 1$.
        Let the indicator function, which is 1 if proposition $P$ is true and 0 otherwise, be denoted as $\ind{P}$.
        The proof is presented in \Cref{subsec:proof:lemma:diff_f}.
        \begin{lemma}[Distribution of the difference of two averages]\label{lemma:diff_f}
            Let $f(x; {\bm \epsilon})$ satisfy \Cref{asm:function} and
            ${\bm \epsilon} = \sbra{\epsilon_1, \ldots, \epsilon_M}^\mathrm{T}$ satisfy \Cref{asm:noise}. 
            In addition, let  
            ${\bm \epsilon}_{i, k} (i=1, 2$ and $k = 1, \ldots, K )$ be i.i.d.\ copies of ${\bm \epsilon}$,
            $\Gamma = \mathrm{diag}(\gamma_1, \ldots, \gamma_M)$, and $\Delta = [\delta_1, \ldots, \delta_M]^\mathrm{T}$.
            Define
             \begin{align}
                \beta'(x) &= \frac{\sum_m^M\beta_m\cdot\mathrm{sign}(g_m(x))\cdot|g_m(x)\gamma_m|^\alpha}{\gamma'(x)^\alpha}
                ,
                \\
                \gamma'(x) &= \norm{\Gamma\nabla_{\bm \epsilon} f(x)}_\alpha
                ,
                \\
                \delta'(x) &= -\frac{2}{\pi}\cdot\sum_m^M \beta_m\gamma_m g_m(x)\cdot\log|g_m(x)|
                ,
            \end{align}
            and
            \begin{align}
                \beta''(x_1, x_2) &=
                \frac{\beta'(x_1)\gamma'(x_1)^\alpha - \beta'(x_2)\gamma'(x_2)^\alpha}{\gamma'(x_1)^\alpha + \gamma'(x_2)^\alpha}
                ,
                \\
                \gamma''(x_1, x_2) &= (\gamma'(x_1)^\alpha + \gamma'(x_2)^\alpha)^{\frac{1}{\alpha}}
                ,
                \\
                \delta''(x_1, x_2) &= \delta'(x_1) - \delta'(x_2)
                .
            \end{align}
            It follows that
            \begin{multline}
                \frac{1}{K}\sum_k^K f(x_1; {\bm \epsilon}_{1, k}) 
                - \frac{1}{K}\sum_k^K f(x_2; {\bm \epsilon}_{2, k})
                \\
                \sim
                {\bm S}\Big(\alpha, 
                \beta''(x_1, x_2),
                K^{\frac{1}{\alpha}-1}\cdot\gamma''(x_1, x_2),
                \\
                f(x_1; \Delta) - f(x_2; \Delta) + \delta''(x_1, x_2)\cdot\ind{\alpha=1}
                \Big)
                .
            \end{multline}
        \end{lemma}

        The following main result allows to evaluate the OEP for $\widehat\eta_\epsilon^\mathrm{AVE}(x_1, x_2)$.
        The proof is presented in \Cref{subsec:proof:theo:oep_stabledist}.
        \begin{theorem}[OEP over stable distribution]\label{theo:oep_stabledist}
            Let $f(x; {\bm \epsilon})$ satisfy \Cref{asm:function} and
            ${\bm \epsilon} = \sbra{\epsilon_1, \ldots, \epsilon_M}^\mathrm{T}$ satisfy \Cref{asm:noise}. In addition, let ${\bm \epsilon}_{i, k} (i=1, 2$ and $k = 1, \ldots, K )$ be i.i.d.\ copies of ${\bm \epsilon}$, $\Gamma = \mathrm{diag}(\gamma_1, \ldots, \gamma_M)$, and $\Delta = [\delta_1, \ldots, \delta_M]^\mathrm{T}$.
            Define
            \begin{multline}
                \epsilon^\mathrm{AVE} \sim 
                {\bm S}\Bigg(
                    \alpha, \beta''(x_1, x_2), 
                    1,
                    \\
                    \left(
                    \frac{\delta''(x_1, x_2)}{\gamma''(x_1, x_2)} + \frac{2}{\pi}\beta''(x_1, x_2)\log(\gamma''(x_1, x_2))
                    \right)\cdot\ind{\alpha = 1}
                    \Bigg)
                .
            \end{multline}
            For any pair $(x_1, x_2)$ with $\eta_\Delta(x_1, x_2) \neq 0$, we have 
            \begin{multline}
                \Pr_{\bm \epsilon}\left[
                    \widehat\eta_\epsilon^\mathrm{AVE}(x_1, x_2)
                    = \eta_\Delta(x_1, x_2)
                \right]
                \\
                =
                \Pr_{\bm \epsilon}\left[
                    -\eta_\Delta(x_1, x_2)\cdot
                    \epsilon^\mathrm{AVE} <
                    \frac{K^{1-\frac{1}{\alpha}}\cdot|f(x_1; \Delta) - f(x_2; \Delta)|}{\gamma''(x_1, x_2)}
                \right]
                \label{eq:oep}
                .
            \end{multline}
            If $\eta_\Delta(x_1, x_2) = 0$, then $\Pr_{\bm \epsilon}\left[
                    \widehat\eta_\epsilon^\mathrm{AVE}(x_1, x_2)
                    = \eta_\Delta(x_1, x_2)
                \right] = 0$.
        \end{theorem}
        
        The implications of \Cref{theo:oep_stabledist} are as follows, 
        noting that the distribution of $\epsilon^\mathrm{AVE}$ is independent of $K$.
        \begin{itemize}
            \item For $1 < \alpha \leq 2$, the OEP approaches $1$ as $K$ increases because the upper bound in the probability increases to infinity. However, the increase rate of the OEP with $K$ is slower for a smaller $\alpha$. 
            \item For $\alpha = 1$, explicit averaging has no effect on the OEP. A high $K$ simply requires more $f$ queries.
            \item For $0 < \alpha < 1$, the OEP decreases as $K$ increases because the upper bound in the probability approaches 0.
        \end{itemize}
        The success of explicit averaging for $1 < \alpha \leq 2$ can be understood considering the law of large numbers. Average $\frac{1}{K}\sum_k^K f(x; {\bm \epsilon}_{k})$ converges to expected value $\E_{{\bm \epsilon}}[f(x; {\bm \epsilon})] = f(x; \Delta)$ if it exists as a finite value. 
        If the expected value does not exist ($0 < \alpha \leq 1$), explicit averaging has a negative impact on the OEP, unless $\alpha = 1$, for which no impact occurs other than the increased cost of more $f$ queries for a greater $K$. 

        To further clarify the OEP, we assume that the distribution is symmetric, that is, $\beta_m=0$ for all $m = 1, \dots, M$.
        In this case, both $\beta''(x_1, x_2)$ and $\delta''(x_1, x_2)$ in \Cref{lemma:diff_f} are zero for any pair $(x_1, x_2)$, and the expression of the OEP is simplified.
        The proof is presented in \Cref{subsec:proof:cor:oep_stabledist_sym}.
        \begin{corollary}[OEP over ${\bf S}(\alpha, 0, \gamma, \delta)$]\label{cor:oep_stabledist_sym}
            Assume $\beta_m = 0$ for all $m=1, \ldots, M$.
            Let $\phi_\alpha(\cdot)$ be the cumulative distribution function of ${\bf S}(\alpha, 0, 1, 0)$.
            For any pair $(x_1, x_2)$ with $\eta_\Delta(x_1, x_2) \neq 0$, we have 
            \begin{multline}
                \Pr_{\bm \epsilon}\left[
                    \widehat\eta_\epsilon^\mathrm{AVE}(x_1, x_2)
                    = \eta_\Delta(x_1, x_2)
                \right]
                \\
                =
                \phi_\alpha\left(
                    \frac
                    {K^{1-\frac{1}{\alpha}}|f(x_1; \Delta) - f(x_2; \Delta)|}
                    {\left(\|\Gamma\nabla_{\bm \epsilon} f(x_1)\|_\alpha^\alpha+\|\Gamma\nabla_{\bm \epsilon} f(x_2)\|_\alpha^\alpha\right)^{\frac{1}{\alpha}}}
                \right)
                .
            \end{multline}
            If $\eta_\Delta(x_1, x_2) \neq 0$, then $\Pr_{\bm \epsilon}\left[
                    \widehat\eta_\epsilon^\mathrm{AVE}(x_1, x_2)
                    = \eta_\Delta(x_1, x_2)
                \right] = 0$.
        \end{corollary}
        Because $\phi_{\alpha}$ is independent of $(x_1, x_2)$ (note that the distribution of $\epsilon^\mathrm{AVE}$ may depend on $(x_1, x_2)$ when $\beta_m \neq 0$), we discuss the impact of $(x_1, x_2)$ on the OEP. 
        First, the OEP monotonically increases as the difference between the ground-truth objective values, $\abs{f(x_1; \Delta) - f(x_2; \Delta)}$, increases. 
        Second, the OEP is higher if $\|\Gamma\nabla_{\bm \epsilon} f(x)\|_\alpha$ is lower for both $x_1$ and $x_2$, where $\|\Gamma\nabla_{\bm \epsilon} f(x)\|_\alpha$ determines the scaling factor for the randomness of the objective function. Indeed, if $\alpha = 2$, the scaling factor is the standard deviation of the objective function value, $f(x; {\bm \epsilon})$. The OEP is higher if the ground-truth objective function difference (i.e., signal) relative to noise scaling is greater. 

\section{Sign Averaging}\label{sec:sign_averaging}

    A negative explicit averaging result occurs when the noise distribution has stability parameter $\alpha\in(0, 1]$.
    This is because expected value $\E_{{\bm \epsilon}}[f(x; {\bm \epsilon})]$ does not exist, and the average value does not converge. 
    Hence, we propose an alternative strategy for overcoming these limitations under different assumptions.
    The proposed \emph{sign averaging} takes the average sign of the differences in the objective function values of two candidate solutions. 
    We theoretically reveal that sign averaging allows to estimate the order of the medians of $f(x_1; {\bm \epsilon})$ and $f(x_2; {\bm \epsilon})$ even when expected values $\E[f(x_1; {\bm \epsilon})]$ and $\E[f(x_2; {\bm \epsilon})]$ do not exist.
    We also propose an approach for incorporating sign averaging into a comparison-based multipoint search algorithm.

    \subsection{Problem Statement}\label{subsec:sa_asm}

        We describe the optimization objective and assumption on $f$ as follows. Instead of assuming the linearity of $f$ with respect to ${\bm \epsilon}$ (\Cref{asm:function}) and a stable distribution of ${\bm \epsilon}$ (\Cref{asm:noise}), we make the following technical assumption.
        The median difference between noisy objective function values $f(x; {\bm \epsilon}_x)$ and $f(y; {\bm \epsilon}_y)$ is assumed to be the difference between their medians.
        \begin{assumption}[Additivity of median]\label{asm:symmetric}
            Let ${\bm \epsilon}\in\R^M$ be a random vector and $f(x; {\bm \epsilon}): \R^D\times\R^M \to \R$.
            In addition, let $\phi_{f}(t; x) = \Pr_{\bm \epsilon}[f(x; {\bm \epsilon}) \leq t]$ be the cumulative distribution function of $f(x; {\bm \epsilon})$.
            Let $m_f(x)$ be the center of the median of $f(x; {\bm \epsilon})$ defined as
            \begin{equation}
                m_f(x) = \frac{\inf_t\{\phi_f(t; x) > 1/2\} + \sup_t\{\phi_f(t; x) < 1/2\}}{2}.
            \end{equation}
            Let $m_f(x, y)$ be the center of the median of $f(x; {\bm \epsilon}) - f(y; {\bm \epsilon}')$, where ${\bm \epsilon}'$ is an i.i.d.\ copy of $\epsilon$. 
            Then, $m_f(x, y) = m_f(x) - m_f(y)$. 
        \end{assumption}

        In general, the median does not satisfy additivity $m_f(x, y) = m_f(x) - m_f(y)$. 
        However, \Cref{asm:symmetric} holds, for example, when the distribution of $f(x; {\bm \epsilon})$ is symmetric for all $x \in \R^D$, as formally stated below. The proof is provided in \Cref{apdx:prop:symmetric}.
        
        \begin{proposition}[Sufficient condition for \Cref{asm:symmetric}]\label{prop:symmetric}
        Suppose that the distribution of $f(x; {\bm \epsilon})$ is symmetric around $m_f(x)$ for any $x \in \R^D$. Then, \Cref{asm:symmetric} holds.            
        \end{proposition}

        \new{
        \Cref{asm:symmetric} holds also when the distribution of $f(x; {\bm \epsilon}) - m_f(x)$ is identical for all $x \in \R^D$. 
        This case includes an often assumed situation, namely the additive noise situation, where the noisy objective function $f(x; {\bm \epsilon}) = f(x) + \xi(\bm{\epsilon})$ is the sum of the ground truth function $f(x)$ and the noise $\xi(\bm{\epsilon})$ that is independent of $x$. 
        It is formally stated below. The proof is provided in \Cref{apdx:prop:identity}.

        \begin{proposition}[Sufficient condition for \Cref{asm:symmetric}]\label{prop:identity}
            Suppose that the distribution of $f(x; {\bm \epsilon}) - m_f(x)$ is identical for all $x \in \R^D$. Then, \Cref{asm:symmetric} holds.  
        \end{proposition}
        }

        The median is not necessarily unique, but its center is, thus hindering analysis. 
        To mitigate this technical difficulty, we assume the following properties.
        
        \begin{assumption}[Regularity of $\phi_f$]\label{asm:uniqueness}
            For any $x \in \R^D$, the cumulative distribution function, $\phi_f(t; x)$, is differentiable at $t = m_f(x)$, and its derivative (i.e., probability density function) is strictly positive at $t = m_f(x)$.
        \end{assumption}

\new{\Cref{asm:uniqueness} is not very restrictive. If \Cref{asm:uniqueness} is not satisfied directly by $f(x; \epsilon)$, it is satisfied by injecting an arbitrarily small noise to each query, as formally stated below. The proof is  provided in \Cref{apdx:prop:uniqueness}.
\begin{proposition}[Sufficient condition for \Cref{asm:uniqueness}]\label{prop:uniqueness}
If we replace each query with $x \mapsto f(x; \bm{\epsilon}) + \xi$, where $\xi$ is an absolutely continuous random variable whose probability density is continuous and strictly positive over $\R$, and $\xi$ and $\bm{\epsilon}$ are independent, then \Cref{asm:uniqueness} holds.
\end{proposition}}
        
        These two assumptions are satisfied for various symmetric distributions.
        For example, under \Cref{asm:function} and \Cref{asm:noise}, the two assumptions above are satisfied when $\beta_m = 0$ for all $m = 1, \dots, M$ (see Theorem 1.9 and Lemma 1.10 of \cite{nolan2020stable}).
        Moreover, these assumptions hold when neither \Cref{asm:function} nor \Cref{asm:noise} is satisfied.
        Some representative probability distributions satisfying \Cref{asm:symmetric} and \Cref{asm:uniqueness} are the continuous uniform, Student's t, Laplace, Bates, and arcsine distributions as well as part of the beta distribution.

        In the following analysis, optimization index $\mathcal{A}[f(x; {\bm \epsilon})]$ for minimization is the center of median $m_f(x)$ of the objective function. 
        In contrast to the optimization index considered in \Cref{sec:oep}, $f(x; \Delta)$, which requires \Cref{asm:noise}, and the expected value, $\E_{\bm \epsilon}[f(x; {\bm \epsilon})]$, which requires the existence of the expected value for all $x \in \R^D$, the median always exists, and its center is uniquely determined. 
        Although these concepts differ, they are related. 
        Under \Cref{asm:function} and \Cref{asm:noise} with $\beta_m = 0$ for all $m = 1, \dots, M$,
        we have $f(x; \Delta) = m_f(x) = \E_{{\bm \epsilon}}[f(x; {\bm \epsilon})]$ for all $\alpha \in (1, 2]$ and $f(x; \Delta) = m_f(x)$ for all $\alpha \in (0, 1]$, whereas $\E_{\bm \epsilon}[f(x; {\bm \epsilon})]$ does not exist.

    \subsection{OEP}\label{subsec:sa}

        Like in \Cref{sec:oep}, we investigate the OEP on the median as follows.
        \begin{definition}[OEP on median]\label{def:oep_sa}
            Let $\widehat\eta_{{\bm \epsilon}}(x_1, x_2)\in \{-1, 0, 1\}$ be an estimate of 
            \begin{equation}
            \eta_\mathrm{med}(x_1, x_2) := \mathrm{sign}\rbra*{
            m_f(x_1) - m_f(x_2)
            }
            .
            \label{eq:eta}
            \end{equation}
            The \emph{OEP} on the median is given by
            \begin{equation}
                \Pr_{\bm \epsilon}[\widehat\eta_{\bm \epsilon} (x_1, x_2) = \eta_\mathrm{med}(x_1, x_2)]
                .
            \end{equation}
        \end{definition}

        We define the order estimate based on sign averaging as
        \begin{equation}
            \widehat\eta_{\bm \epsilon}^\mathrm{SA}(x_1, x_2)
            :=
            \mathrm{sign}\left(\frac{1}{K}\sum_k^K\mathrm{sign}\left(f(x_1; {\bm \epsilon}_{1, k}) - f(x_2; {\bm \epsilon}_{2, k})\right)\right)
            .
            \label{eq:eta_sa}
        \end{equation}
        Consider its expected value:
        \begin{equation}
        \E\left[\mathrm{sign}\left(f(x_1; {\bm \epsilon}_{1}) - f(x_2; {\bm \epsilon}_{2})\right) \right].
        \end{equation}
        It exists as a finite value, and its sample average must converge to the expected value as $K$ increases.
        Moreover, the following proposition guarantees that the sign of expected value $\E\left[\mathrm{sign}\left(f(x_1; {\bm \epsilon}_{1}) - f(x_2; {\bm \epsilon}_{2})\right) \right]$ always agrees with $\eta_\mathrm{med}(x_1, x_2)$. 
        The proof is provided in \Cref{subsec:proof:prop:SA_symmetric}.
        \begin{proposition}[Expectation of $\mathrm{sign}\rbra*{f(x_1; {\bm \epsilon}_1) - f(x_2; {\bm \epsilon}_2)}$]\label{prop:SA_symmetric}
            Let random vector ${\bm \epsilon}\in\R^M$ and objective function $f(x; {\bm \epsilon})$ satisfy \Cref{asm:symmetric}.
            Define
            \begin{equation}
                \epsilon^f(x_1, x_2) := f(x_1; {\bm \epsilon}_1) - f(x_2; {\bm \epsilon}_2)
                ,
            \end{equation}
            where ${\bm \epsilon}_1, {\bm \epsilon}_2$ are i.i.d.\ copies of ${\bm \epsilon}$.
            For any $(x_1, x_2)$, we have
            \begin{equation}
            \E\left[\mathrm{sign}(\epsilon^f(x_1, x_2))\right] \cdot \eta_\mathrm{med}(x_1, x_2) \geq 0.
            \label{eq:signproduct}
            \end{equation}
            Moreover, if \Cref{asm:uniqueness} is satisfied, we have the following relation for any $(x_1, x_2)$ satisfying $\eta_\mathrm{med}(x_1, x_2) \neq 0$:
            \begin{equation}
            \E\left[\mathrm{sign}(\epsilon^f(x_1, x_2))\right] \cdot \eta_\mathrm{med}(x_1, x_2) > 0.
            \label{eq:signequality}
            \end{equation}
        \end{proposition}

        The following OEP bound with $\widehat\eta_{\bm \epsilon}^\mathrm{SA}(x_1, x_2)$ guarantees that the probability of $\E\left[\mathrm{sign}(\epsilon^f(x_1, x_2))\right]\cdot\eta_\mathrm{med}(x_1, x_2) > 0$ being satisfied approaches $1$ as $K$ increases if $\eta_\mathrm{med}(x_1, x_2) \neq 0$.
        The proof is presented in \Cref{subsec:proof:theo:oep_stabledist_sign}.
        \begin{theorem}[OEP with sign averaging]\label{theo:oep_stabledist_sign}
            Let random vector ${\bm \epsilon}\in\R^M$ and objective function $f(x; {\bm \epsilon})$ satisfy \Cref{asm:symmetric}.
            In addition, let ${\bm \epsilon}_{i, k} (i=1, 2$ and $k = 1, \ldots, K )$ be an i.i.d.\ copy of ${\bm \epsilon}$.
            If $\eta_\mathrm{med}(x_1, x_2)\neq 0$,
            it holds that
            \begin{multline}
                \Pr_{\bm \epsilon}\left[
                \widehat\eta_{\bm \epsilon}^\mathrm{SA}(x_1, x_2)
                = \eta_\mathrm{med}(x_1, x_2)
                \right]
                \\
                \geq
                1-
                \exp\rbra*{-\frac{K\cdot\rbra*{\E\left[\mathrm{sign}\left(\epsilon^f(x_1, x_2)\right)\right]}^2}{2}}
                .
                \label{eq:positive_eps_sa}
            \end{multline}
        \end{theorem}

        The implications of \Cref{theo:oep_stabledist_sign} are as follows. 
        First, we observe that the OEP lower bound increases with increasing $K$. 
        Because \Cref{asm:symmetric} and \Cref{asm:uniqueness} are satisfied by \Cref{asm:function} and \Cref{asm:noise} with $\beta_m = 0$ (for all $m = 1, \dots, M$), the result indicates that sign averaging is effective even when $\alpha \leq 1$, whereas explicit averaging fails.
        Hence, sign averaging is a valid alternative to explicit averaging in comparison-based algorithms.
        Second, the OEP lower bound with respect to pair $(x_1, x_2)$ is higher for a pair with larger $\E\left[\mathrm{sign}\left(\epsilon^f(x_1, x_2)\right)\right]^2$. This expectation can be rewritten as 
        \begin{multline}
         \E\left[\mathrm{sign}\left(\epsilon^f(x_1, x_2)\right)\right]
         \\
         =
        \Pr[f(x_1; \epsilon_1) < f(x_2; \epsilon_2)] - \Pr[f(x_1; \epsilon_1) > f(x_2; \epsilon_2)]
         .\label{eq:esign}
        \end{multline}
        Hence, a higher probability of a solution being better than or smaller than the other one increases the OEP lower bound.
        
        Considering \Cref{prop:SA_symmetric}, \Cref{asm:uniqueness} guarantees 
        \eqref{eq:signequality} and the OEP lower bound with $\eta_{\bm \epsilon}^\mathrm{SA}(x_1, x_2)$ can arbitrarily approach 1 by increasing $K$.
        Sample size $K$ for sign averaging to obtain an OEP above $p$ is $O\rbra*{\log\rbra*{\frac{1}{1-p}}}$, as formally stated below.
        \begin{corollary}[Sufficient sample size $K$ to estimate OEP]\label{cor:K_to_p}
        Suppose that \Cref{asm:symmetric} and \Cref{asm:uniqueness} hold. Then, for any pair $(x_1, x_2)$ satisfying $\eta_\mathrm{med}(x_1, x_2) \neq 0$ and for any $p \in (0, 1)$, we have 
        $\Pr_{\bm \epsilon}\left[
                \widehat\eta_{\bm \epsilon}^\mathrm{SA}(x_1, x_2)
                = \eta_\mathrm{med}(x_1, x_2),
                \right]
                \geq p$ if 
            \begin{equation}
                K \geq \frac{2}{\E\left[\mathrm{sign}\left(\epsilon^f(x_1, x_2)\right)\right]^2}\cdot\log\rbra*{\frac{1}{1-p}}
                .
            \end{equation}
        \end{corollary}

\subsection{Weighting Based on Sign Averaging}\label{subsec:sacma}

We propose a mechanism for incorporating sign averaging into comparison-based optimization algorithms.

A comparison-based optimization algorithm typically has a set of candidate solutions, $\{x_i\}_{i=1}^{\lambda}$, which is often called the population. If objective function $f$ is deterministic, the population is sorted in ascending order of $f$, that is, $f(x_{1:\lambda}) \leq \dots \leq f(x_{\lambda:\lambda})$, where $i:\lambda$ denotes the index of the $i$-th best point among $\lambda$ points. Ties (i.e., $f(x_i) = f(x_j)$) are usually ignored because they are unlikely during continuous optimization. Based on the rankings, predefined weights are assigned to the candidate solutions. Let $w_1, \dots, w_\lambda$ be predefined weights. Typically, $w_1 \geq \dots \geq w_\lambda$. Then, $x_{i:\lambda}$ receives $w_i$. The internal information of the optimization algorithm is updated using $\{(x_{i:\lambda}, w_i)\}_{i=1}^{\lambda}$. 
For example, in CMA-ES \cite{hansen2001completely,hansen2003reducing,hansen2006cma}, the internal information comprises the parameters of the normal distribution and evolution paths. The predefined weights can differ for updates of different internal parameters, like for the latest CMA-ES~\cite{akimoto2020diagonal} (i.e., when active covariance matrix adaptation \cite{jastrebski2006improving} is used).

One technical difficulty in incorporating sign averaging into comparison-based optimization algorithms is that $\widehat\eta_{\bm \epsilon}^\mathrm{SA}$ is not necessarily transitive. That is, for three points $x_1, x_2, x_3$, $\widehat\eta_{\bm \epsilon}^\mathrm{SA}(x_1, x_2) < 0$, and $\widehat\eta_{\bm \epsilon}^\mathrm{SA}(x_2, x_3) < 0$, $\widehat\eta_{\bm \epsilon}^\mathrm{SA}(x_1, x_3) > 0$ may occur. In such case, the rankings of $x_1, x_2, x_3$ cannot be determined. 
If this occurs, weighting cannot be directly applied. 
To sort population $\{x_i\}_{i=1}^{\lambda}$ using $\widehat\eta_{\bm \epsilon}^\mathrm{SA}$, the score of each point $x_i$ is defined as the number of points that are estimated to be better than $x_i$, namely, 
\begin{equation}
    R\left(x_i; \{x_j\}_{j=1}^\lambda\right) := 
    \sum_{j=1}^\lambda \ind{\widehat\eta_{\bm \epsilon}^\mathrm{SA}(x_j, x_i) \leq 0}
    ,
    \label{eq:R}
\end{equation}
and the population can be sorted in ascending order of $R$. 
If the population rankings are uniquely determined by $\widehat\eta_{\bm \epsilon}^\mathrm{SA}$ (i.e., $\widehat\eta_{\bm \epsilon}^\mathrm{SA}$ is transitive on $\{x_j\}_{j=1}^{\lambda}$), \eqref{eq:R} is the ranking of $x_i$. Therefore, it may be considered as an extension of the ranking mechanism using $\widehat\eta_{\bm \epsilon}^\mathrm{SA}$.
However, ties may occur and should be handled carefully to avoid undermining the performance of the resulting algorithm. 

In \cite{akimoto2022ode}, a weighting scheme was proposed to treat ties with mathematical rigor for deterministic objective functions. 
Weighting introduced in \cite{akimoto2022ode} is defined as follows.
For each $x_i \in \{x_j\}_{j=1}^\lambda$, the numbers of strictly better and weakly better points among $\{x_j\}_{j=1}^\lambda$ are counted as follows:
\begin{align}
    &r^<_i := 
    \sum_{j = 1}^\lambda \ind{f(x_j) < f(x_i)}
    ,\label{eq:rlt}
    \\
    &r^\leq_i := \sum_{j=1}^\lambda \ind{f(x_i) \leq f(x_j)}
    .\label{eq:rleq}
\end{align}
Based on these quantities, a weight is assigned to every $x_i$ as follows:
\begin{equation}
    \bar w_i
    := 
    \frac{1}{r^\leq_i-r^<_i}
    \cdot\sum_{j = r^<_i + 1}^{r^\leq_i} w_j
    .
    \label{eq:sa_weight}
\end{equation}
If there is no tie, $r_i^{\leq} = r_i^{<} + 1$ for all $i=1,\dots,\lambda$ and $r_i^{\leq}$ agrees with the conventional $x_i$ ranking. The assigned weights are equivalent to the aforementioned conventional weighting.
If ties exist, the same weight is assigned to the tie points. Thus, we can guarantee that the sum of assigned weights is the same as the sum of the predefined weights, that is, $\sum_{i=1}^{\lambda} \bar{w}_i = \sum_{i=1}^{\lambda} w_i$.

We incorporate averaging into weighting in \eqref{eq:sa_weight} by simply replacing $f(\cdot)$ with $R\left(\cdot; \{x_j\}_{j=1}^\lambda\right)$ in \eqref{eq:rlt} and \eqref{eq:rleq}. If $f$ is deterministic and no ties exist among $f(x_1), \dots, f(x_\lambda)$, the assigned weights are the same as those obtained by conventional weighting. If $f$ is stochastic but the rankings of the population are uniquely determined by $\widehat\eta_{\bm \epsilon}^\mathrm{SA}$ (i.e., $\widehat\eta_{\bm \epsilon}^\mathrm{SA}$ is transitive on $\{x_j\}_{j=1}^{\lambda}$), the weights computed by the proposed approach are the same as those assigned by conventional weighting with $\widehat\eta_{\bm \epsilon}^\mathrm{SA}$ for comparison. 
Moreover, according to \Cref{theo:oep_stabledist_sign}, the weights assigned by the proposed approach agree with conventional weighting with $\eta_\mathrm{med}$ for comparison in limit $K \to \infty$ if no tie exists.

\section{Experiments on OEP}\label{sec:experiment}
 
        \begin{figure*}[t]
            \centering
            \includegraphics[width=18cm]{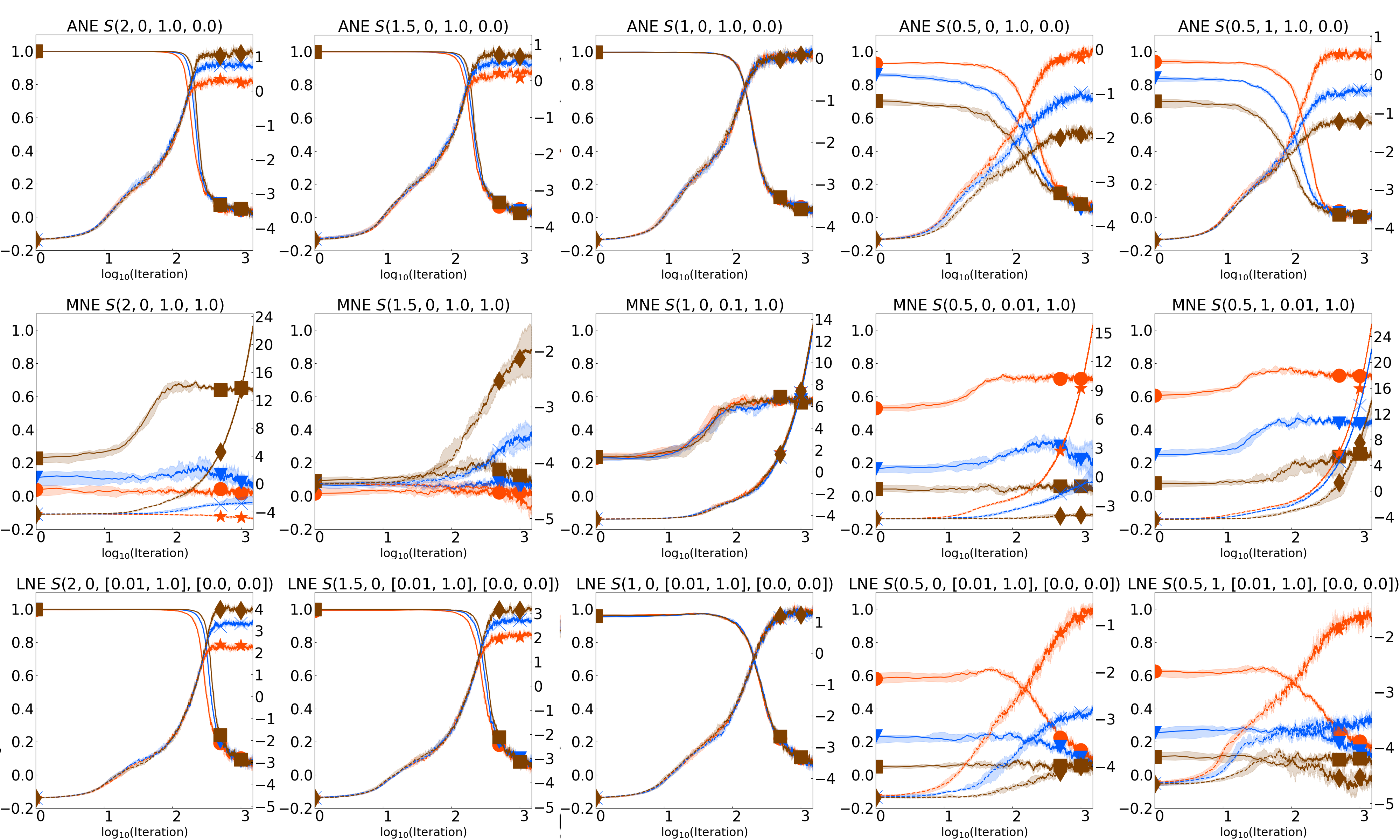}
            \caption{Results of 10 runs of CMA-ES with explicit averaging. 
            The solid lines indicate the median of the moving average of Tau-b (left axis) with sample sizes $K=1$ ($\bullet$), $K=10$ ($\blacktriangledown$), and $K=50$ ($\blacksquare$). 
            The span of the moving average is $10$.
            The dashed lines indicate $-\log(f(m_t; \Delta))$ (right axis) with sample sizes $K=1$ ($\star$), $K=10$ ($\times$), and $K=50$ ($\blacklozenge$).
            The top and bottom of the band correspond to the 75\% and 25\% values among trials.
            The medians are taken over $10$ trials. (ANE, additive-noise ellipsoid; LNE, linear-noise ellipsoid; MNE, multiplicative-noise ellipsoid)
            }
            \label{fig:averaging}
        \end{figure*}     

    \begin{figure*}[t]
        \centering
        \includegraphics[width=18cm]{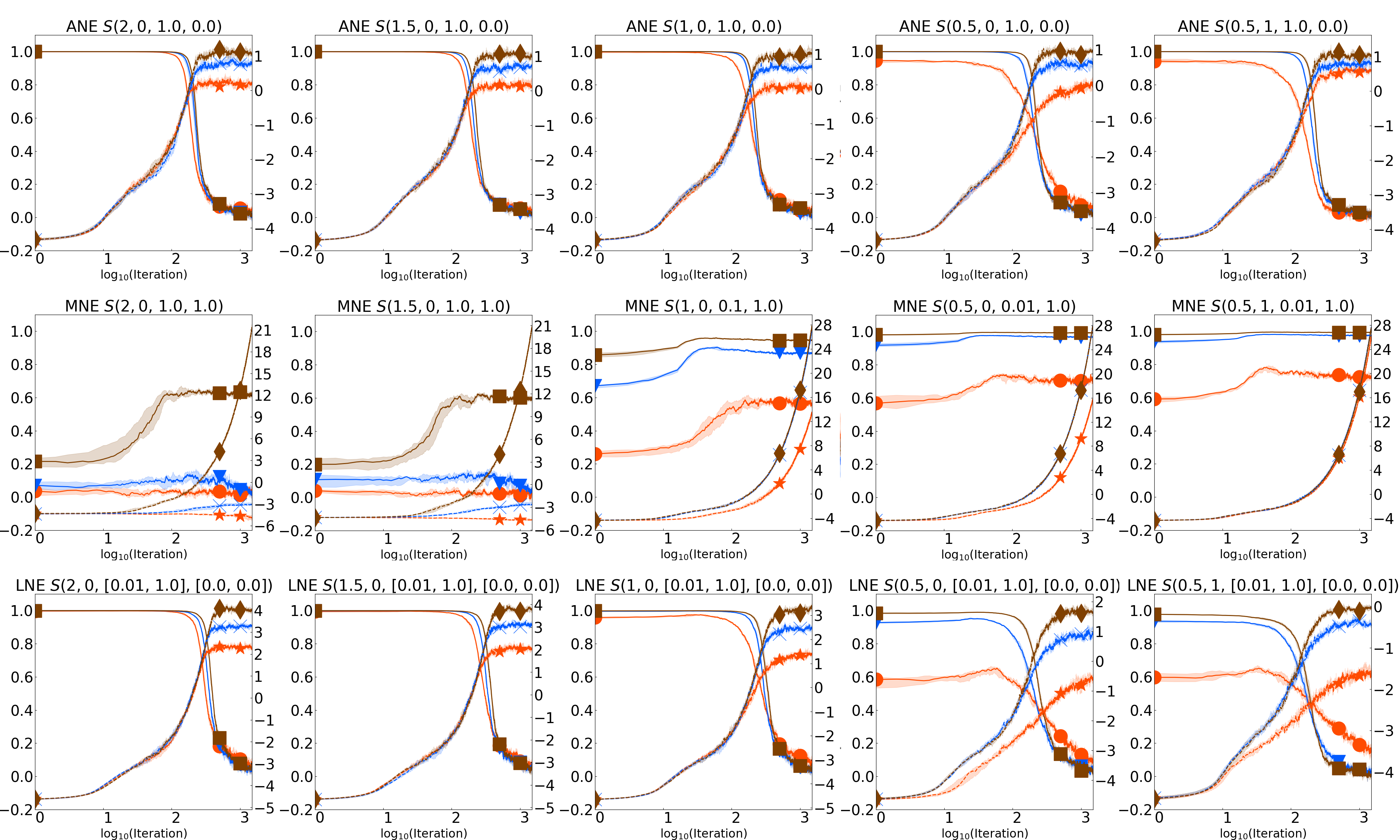}
        \caption{Results of 10 runs of CMA-ES with sign averaging. The solid lines indicate the median of the moving average of Tau-b (left axis) with sample sizes $K=1$ ($\bullet$), $K=10$ ($\blacktriangledown$), and $K=50$ ($\blacksquare$). 
        The dashed lines indicate $-\log(f(m_t; \Delta))$ (right axis) with sample sizes $K=1$ ($\star$), $K=10$ ($\times$), and $K=50$ ($\blacklozenge$).}
        \label{fig:sign_averaging}
    \end{figure*}     

    First, we incorporated explicit averaging into $(\mu/\mu_w, \lambda)$-CMA-ES~\cite{hansen2001completely, hansen2003reducing} as a representative comparison-based optimization algorithm.
    The negative effect of averaging was confirmed when the value of stability parameter $\alpha$ was less than $1$ and positive for $\alpha \in (1, 2]$.
    Under the same experimental settings, explicit averaging was replaced by sign averaging using the weighting scheme discussed in \Cref{subsec:sacma}.
    Sign averaging contributed to the CMA-ES accuracy for approximating the optimum, even for $\alpha\in(0, 1]$ at which conventional averaging failed.

    In the abovementioned experiments, two optimization indices were obtained per iteration.
    The first index was the objective function value with location parameter $f(m_t, \Delta)$, which was assumed to be the ground-truth $f$ value.
    Again, this index agreed with both the median of $f(m_t, \epsilon)$ on symmetric stable noise and the mean if it was finite.
    The second index was the Kendall rank correlation coefficient, Tau-b, which is defined on pairs of numerical values as the difference between the number of concordant pairs and the number of discordant pairs divided by the number of pairs that are not tied~\cite{agresti2010analysis}.
    In each experiment, Tau-b was applied to a set of pairs of 
    the estimated ranking of the ground-truth $f$ value, $f(x; \Delta)$, and the proper ranking in the population at each iteration was calculated.
    Tau-b is 1 if noise handling attains perfect ranking estimation and $-$1 otherwise, while it is 0 if the estimated and ground-truth rankings are uncorrelated.

    \paragraph{Test functions}
        We define three test functions that satisfy \Cref{asm:function}. 
        1) Additive-noise ellipsoid $f_\mathrm{add}(x; {\bm \epsilon})$ is defined as $f_\mathrm{add}(x; {\bm \epsilon}) = x^\mathrm{T}Hx + \epsilon$, where $H\in\R^{D\times D}$ is a positive definite diagonal matrix with the $i$-th eigenvalue being $10^{2(i-1)/(D-1)}$, 
         and $\epsilon\in\R$ is a random variable following a stable distribution. The values of these parameters are listed in the titles of the corresponding figures.
        
        2) Multiplicative-noise ellipsoid $f_\mathrm{mul}(x; {\bm \epsilon})$ is defined as $f_\mathrm{mul}(x; {\bm \epsilon}) = x^\mathrm{T}Hx\cdot \epsilon$, where $H\in\R^{D\times D}$ is identical to that of the additive-noise ellipsoid and $\epsilon\in\R$ is stable noise.

        3) Linear-noise ellipsoid $f_\mathrm{LNE}(x; {\bm \epsilon})$ is defined as $f_\mathrm{LNE}(x; {\bm \epsilon}) = x^\mathrm{T}Hx + {\bm \epsilon}^\mathrm{T} x$, 
        where $H$ is identical to that of the additive noise ellipsoid
        and ${\bm \epsilon}\in\R^M$ is a random vector whose $m$-th element is stable noise ${\bf S}(\alpha, \beta, \gamma_m, \delta_m)$.
        We set $\gamma_m = 10^{a+(m-1)(b-a)/(D-1)}$ and 
        $\delta_m = 0$. The values of $a, b$ are specified as ${\bm S}(\alpha, \beta, [10^a, 10^b], [0, 0])$ in the titles of the corresponding figures.

        For the three defined test functions, the ground-truth objective function value is $f(x; \Delta) = x^\mathrm{T}Hx$ as we set $\delta_m = 0$ for all $m =1, \dots, M$.
        
    \paragraph{Parameters of noise ${\bm \epsilon}$}

        We investigated the ability of algorithms to optimize the three test functions under five types of stable noise components: 1) normal distribution ($\alpha=2, \beta$), 2) symmetric stable distribution with $\alpha=1.5, \beta=0$, 3) Cauchy distribution ($\alpha=1, \beta=0$), 4) symmetric stable distribution with $\alpha=0.5, \beta = 0$, and 5) L\'evy distribution ($\alpha =0.5, \beta=1$). 
        The first four distributions were examined to compare the differences between variations in $\alpha$.
        The L\'evy distribution was used to evaluate an asymmetric stable distribution.

    \paragraph{Other experimental settings}
    
        For each test function, the dimension of design variable $x$ was $D=20$. 
        Initial solution $m_0$ was $[10, \ldots, 10]^\mathrm{T}$ for all the runs.
        All the settings of CMA–ES followed the details in~\cite{hansen2006cma}.

    \subsection{Explicit Averaging}

        Averaging was incorporated into CMA-ES. It received $\{\frac{1}{K}\sum_k^K f(x_i; \epsilon_{i, k})\}_i^\lambda$ as the estimated $f$ value of population $\{x_i\}_i^\lambda$.
        \Cref{fig:averaging} shows Tau-b 
        on pairs $\cbra*{\rbra*{\frac{1}{K}f(x_i; \epsilon_{i, k}), f(x_i; \Delta)}}_i^\lambda$ 
        and ground-truth objective function value $-\log(f(m_t; \Delta)$ over iteration $t$ of the CMA-ES with explicit averaging for $K=1, 10, 50$, where $m_t$ is the mean vector of CMA-ES at iteration $t$.
        All the test functions and noise satisfied \Cref{asm:function} and \Cref{asm:noise}.

        In \Cref{fig:averaging}, the implication of \Cref{theo:oep_stabledist} on sample size $K$ was confirmed. 1) For $\alpha\in(1, 2]$, we obtained a lower $f(x; \Delta)$ for a greater $K$ (first and second columns from the left). 2) For $\alpha=1$, averaging had no impact on $\tau$ and $f(x; \Delta)$, and it only increased the number of $f$ calls (third column). 3) For $\alpha\in(0, 1)$, we obtained a lower $f(x; \Delta)$ for a lower $K$ (fourth and fifth columns).

\subsection{Sign Averaging}\label{subsec:experiment_sa}

    Next, we incorporated sign averaging into CMA-ES using the weighting proposed in~\Cref{subsec:sacma}, obtaining sign-averaging CMA-ES.
    We confirmed the effectiveness of sign averaging for problems with noise without a finite mean, for which explicit averaging was unsuccessful.
    For a fair comparison, all the experimental settings and baseline algorithm were as those in \Cref{sec:experiment}, except for explicit averaging, which was replaced by sign averaging.
    Because sign averaging did not directly estimate the $f$ values, Tau-b on the pairs of the estimated rankings of $x_i$ in the population, $R(x_i)$, and the ground-truth value were examined.
    All the test functions with symmetric distributions of the noise used in \Cref{sec:experiment} also satisfied \Cref{asm:symmetric} and \Cref{asm:uniqueness}, and asymmetric noise ($\alpha=0.5, \beta=1$) did not satisfy \Cref{asm:symmetric}.

    \Cref{fig:sign_averaging} shows Kendall's Tau-b on $\rbra*{R(x_i), f(x_i; \Delta)}_i^\lambda$
    and $-\log(f(m_t; \Delta))$ over iteration $t$.
    For $\alpha=2$ (first column in \Cref{fig:sign_averaging}), sign-averaging CMA-ES attained almost the same precision of $-\log(f(m_t; \Delta))$ as that obtained by CMA-ES with explicit averaging (\Cref{fig:averaging}) for all $K$ values in the three test functions.
    For $\alpha=1.5$ (second column in \Cref{fig:sign_averaging}), 
    the performance of sign-averaging CMA-ES remained almost the same as that for $\alpha=2$ regardless of the $K$ value and test function.
    For the other two symmetric distributions (third and fourth columns in \Cref{fig:sign_averaging}), sign-averaging CMA-ES clearly outperformed CMA-ES with explicit averaging on the three test functions as $K$ increased. 
    This finding agreed with the implications of \Cref{theo:oep_stabledist_sign}.
    Furthermore, for asymmetric noise (last column in \Cref{fig:sign_averaging}), while no effect of $K$ appeared in the multiplicative-noise ellipsoid, the performance on index $f(x; \Delta)$ was improved by increasing $K$ in the additive- and linear-noise ellipsoids.

\section{Discussion}\label{sec:discussion}

        \subsection{Invariance under Strictly Increasing Transformation}\label{subsec:invariance}

        Let $G: \R\to\R$ be a strictly increasing transformation such that $G(x)\lesseqgtr G(y)\Leftrightarrow x\lesseqgtr y$.
        The behavior of a comparison-based algorithm is generally invariant to such transformation because the ranking of the objective function values is invariant to $G$.
        However, invariance may break with averaging, which is not invariant to $G$. Hence, the following case may occur: 
        \begin{multline}
            \mathrm{sign}
            \cbra*{
            \frac{1}{K}\sum_k^K G\circ f(x_1; {\bm \epsilon}_1) - 
            \frac{1}{K}\sum_k^K G\circ f(x_2; {\bm \epsilon}_2)
            }
            \\
            \neq
            \mathrm{sign}
            \cbra*{
            \frac{1}{K}\sum_k^K f(x_1; {\bm \epsilon}_1) - 
            \frac{1}{K}\sum_k^K f(x_2; {\bm \epsilon}_2)
            }
            .
        \end{multline}
        On the other hand, sign averaging preserves invariance under $G$. Hence, the following equality always holds:
        \begin{multline}
            \mathrm{sign}
            \cbra*{
            \frac{1}{K}\sum_k^K \mathrm{sign}\rbra*{G\circ f(x_1; {\bm \epsilon}_1) - G\circ f(x_2; {\bm \epsilon}_2)}
            }
            \\
            =
            \mathrm{sign}
            \cbra*{
            \frac{1}{K}\sum_k^K \mathrm{sign}\rbra*{f(x_1; {\bm \epsilon}_1) - f(x_2; {\bm \epsilon}_2)}
            }
            .
        \end{multline}

        \subsection{Asymptotic OEP}
        
        Considering asymptotic normality, we can derive the order of the OEP in $K$ for explicit averaging, sign averaging, and sample median.

        For a global perspective of the derivation, suppose that $\widehat\theta_K$ is an estimate of a parameter $\theta$ with $K$ samples and that $\widehat{\theta}_K$ exhibits asymptotic normality, that is, $Z_K = \frac{\sqrt{K} (\widehat{\theta}_K - \theta)}{\sigma}$ converges in distribution to $\mathcal{N}(0, 1)$ for some $\sigma > 0$. 
        Then, for $\theta \neq 0$, we have
        \begin{subequations}
        \begin{align}
        \Pr\sbra*{\mathrm{sign}(\widehat{\theta}_K) = \mathrm{sign}(\theta)}
        &=
        \Pr\sbra*{\widehat{\theta}_K \cdot \theta > 0}
        \\
        &=
        \Pr\sbra*{ \left( \theta + \frac{\sigma Z_K}{\sqrt{K}} \right)\cdot \theta > 0}
        \\
        &=
        \Pr\sbra*{ Z_K \leq  \frac{\sqrt{K} \abs{\theta} }{\sigma}}
        \\
        &\approx
        \Phi_{\mathcal{N}}\left( \frac{\sqrt{K} \abs{\theta} }{\sigma} \right).
        \end{align}
        \end{subequations}
        Therefore, the probability of correctly estimating the sign of parameter $\theta$ is approximated by the cumulative distribution function of a normal distribution. 

        The OEPs for explicit averaging, sign averaging, and sample median are approximated using asymptotic normality. 
        For simplicity, we assume that the distribution of $f(x; {\bm \epsilon})$ is symmetric and absolutely continuous for any $x$ and that the probability density function of $f(x_1; {\bm \epsilon}_1) - f(x_2; {\bm \epsilon}_2)$ is strictly positive at its unique median. Hence, the following results are obtained:
        \begin{itemize}
            \item (Explicit averaging)
            Let $\widehat\mu_K^\mathrm{AVE}(x_1, x_2) := \frac{1}{K}\sum_k^K f(x_1; {\bm \epsilon}_{1, k})
            - \frac{1}{K}\sum_k^K f(x_2; {\bm \epsilon}_{2, k})$,
            $\mu^\mathrm{AVE}(x_1, x_2) := \E[f(x_1; {\bm \epsilon}_1) - f(x_2; {\bm \epsilon}_2)]$,
            and $\sigma^\mathrm{AVE}(x_1, x_2)^2 := \Var[f(x_1 ; {\bm \epsilon})] + \Var[f(x_2 ; {\bm \epsilon})]$.
            Assume $\mu^{\mathrm{AVE}}(x_1, x_2)<\infty$ and $\sigma^\mathrm{AVE}(x_1, x_2)<\infty$ for any $x$.
            Then, considering the central limit theorem, the asymptotic normality of $\widehat\mu^\mathrm{AVE}(x_1, x_2)$ for $\mu^\mathrm{AVE}(x_1, x_2)$ and $\sigma^\mathrm{AVE}(x_1, x_2)$ holds, and
            for large $K$, the OEP is approximated as
        \begin{multline}
            \Pr\sbra*{
                \mathrm{sign}(\widehat\mu^\mathrm{AVE}(x_1, x_2))
                =
                \mathrm{sign}(\mu^\mathrm{AVE}(x_1, x_2))
            }
            \\
            \approx
            \phi_{\mathcal{N}}\rbra*{\frac{\sqrt{K}\cdot\abs{\mu^\mathrm{AVE}(x_1, x_2)}}{\sigma^\mathrm{AVE}(x_1, x_2)}}
            .
        \end{multline}

        \item
        (Sign averaging)
        Let $\widehat\mu_K^\mathrm{SA}(x_1, x_2) := \frac{1}{K}\sum_k^K\mathrm{sign}(f(x_1; {\bm \epsilon}_{1, k}) - f(x_2; {\bm \epsilon}_{2, k}))$, $\mu^\mathrm{SA} := \E[\mathrm{sign}(f(x_1; {\bm \epsilon}_{1, k}) - f(x_2; {\bm \epsilon}_{2, k}))]$, and $\sigma^\mathrm{SA}(x_1, x_2)^2 := \Var[\mathrm{sign}(f(x_1; {\bm \epsilon}_{1, k}) - f(x_2; {\bm \epsilon}_{2, k}))]$.
        Then, $\mu^\mathrm{SA}(x_1, x_2) < \infty$ and $\sigma^\mathrm{SA}(x_1, x_2)^2 < \infty$.
        Considering the central limit theorem, the asymptotic normality of $\widehat\mu^\mathrm{SA}(x_1, x_2)$ for $\mu^\mathrm{SA}(x_1, x_2)$ and $\sigma^\mathrm{SA}(x_1, x_2)$ holds. 
        For large $K$, the OEP is approximated as 
        \begin{multline}
            \Pr\sbra*{
                \mathrm{sign}(\widehat\mu^\mathrm{SA}(x_1, x_2))
                =
                \mathrm{sign}(\mu^\mathrm{SA}(x_1, x_2))
            }
            \\
            \approx
            \phi_{\mathcal{N}}\rbra*{\frac{\sqrt{K}\cdot\abs{\mu^\mathrm{SA}(x_1, x_2)}}{\sigma^\mathrm{SA}(x_1, x_2)}}
            .
            \label{eq:approx_sa}
        \end{multline}

        \item
        (Sample median)
        Let $\xi(x_1, x_2)$ be the value of the probability density function of $f(x_1; {\bm \epsilon}_1) - f(x_2; {\bm \epsilon}_2)$ at its median.
        In addition, let $\widehat\mu^\mathrm{MED}_K(x_1, x_2)$ be the sample median of $\{f(x_1; {\bm \epsilon}_{1,k}) - f(x_2; {\bm \epsilon}_{2,k})\}_{k=1}^{K}$, $\mu^\mathrm{MED}(x_1, x_2)$ be the population median of $f(x_1; {\bm \epsilon}_1) - f(x_2; {\bm \epsilon}_2)$, and $\sigma^\mathrm{MED}(x_1, x_2) := 
        \frac{1}{2\xi(x_1, x_2)}$.
        Consider the asymptotic normality of the sample median~\cite{casella2021statistical}. 
        For large $K$, the OEP is approximated as
        \begin{multline}
            \Pr\sbra*{
                \mathrm{sign}(\widehat\mu_K^\mathrm{MED}(x_1, x_2))
                =
                \mathrm{sign}(\mu^\mathrm{MED}(x_1, x_2))
            }
            \\
            \approx
            \phi_{\mathcal{N}}\rbra*{\frac{\sqrt{K}\cdot\abs{\mu^\mathrm{MED}(x_1, x_2)}}{\sigma^\mathrm{MED}(x_1, x_2)}}
            .
        \end{multline}
        \end{itemize}
        Note that $\mathrm{sign}(\mu^\mathrm{SA}(x_1, x_2))$ and $\mathrm{sign}(\mu^\mathrm{MED}(x_1, x_2))$ are equivalent for any $x_1, x_2$ considering \Cref{prop:SA_symmetric}. In addition, they are equivalent to $\mathrm{sign}(\mu^\mathrm{AVE}(x_1, x_2))$ if $\mu^\mathrm{AVE}(x_1, x_2)<\infty$.
        Therefore, the three methods estimate the same quantity under the current assumptions.
        Overall, we have the same asymptotic order of OEP with respect to $K$ for the three estimates \new{if the mean of the noise exists as a finite value. 
        We emphasize that the sample median and the sign averaging attains these asymptotic results even if the mean is not defined.}

    \section{Conclusion}\label{sec:conclusion}

        We investigated the effects of explicit averaging on comparison-based search algorithms. In particular, we analyzed the OEP (\Cref{def:oep}) of correctly estimating the order of the ground-truth objective functions of pairs of points. Assuming linearity of the objective function with respect to the noise vector (\Cref{asm:function}) and stably distributed noise (\Cref{asm:noise}), 
        the main results (\Cref{theo:oep_stabledist} and \Cref{cor:oep_stabledist_sym}) reveal the following. 
        1) If the stability parameter of stably distributed noise is $\alpha \in (1, 2]$, increasing sample size $K$ in explicit averaging increases the probability. 2) If $\alpha=1$ (e.g., Cauchy distribution), $K$ has no effect on the probability. 3) If $0<\alpha<1$ (e.g., L\'evy distribution), increasing $K$ decreases the probability.
        Thus, explicit averaging is effective even if the variance of $f(x; {\bm \epsilon})$ does not exist, but it is harmful if the mean of $f(x; {\bm \epsilon})$ does not exist.
        This is experimentally confirmed through optimization using CMA-ES.

        We propose the alternative of sign averaging for estimating the order of the objective function medians on a pair of points even when the objective function mean does not exist.
        This is theoretically demonstrated in \Cref{theo:oep_stabledist_sign} and \Cref{cor:K_to_p} under certain regularity conditions of the noise distribution (\Cref{asm:symmetric} and \Cref{asm:uniqueness}). 
        \new{Any problem satisfying \Cref{asm:function} and \Cref{asm:noise} also satisfies \Cref{asm:symmetric} and \Cref{asm:uniqueness} if the distribution of the noise is symmetric.
        That is, the sign averaging can handle the stable noise for all $\alpha\in(0, 2]$, which includes the cases that the mean does not exists as a finite value, where the averaging fails.}
        Because sign averaging cannot be directly incorporated into common comparison-based algorithms, we propose a versatile weighting scheme.
        The efficiency of sign averaging in this weighting scheme is experimentally confirmed using the CMA-ES.

        \new{In future work, we will address the limitations of this study. 
        First, we will demonstrate how sign-averaging CMA-ES works under diverse noise distributions.
        In this work, the analysis of sign averaging is based on \Cref{asm:symmetric}. It is satisfied when the distribution of the noisy objective function values is symmetric or when the noise distribution is independent of design variable $x$. 
        These assumptions may not hold in some cases. 
        We have not investigated the effect of sign averaging in situations where \Cref{asm:symmetric} does not hold. 
        We will investigate the effect of sign averaging both theoretically and empirically.
        Second, we will theoretically and empirically investigate the effect of the sample median as an alternative noise-handling technique.
        The sample median may be the most natural choice for noise handling when the optimization index is the population median of the objective function.
        However, we have not investigated the effect of the sample median under a finite sample size $K$. 
        We will theoretically and empirically investigate the effect of the sample median and compare it with sign averaging.}

\appendix

\section{Proof}

    \subsection{Proof of \Cref{lemma:diff_f}}\label{subsec:proof:lemma:diff_f}
    \begin{proof}
       First, we prove that the distribution of the sum of $\epsilon_l\sim{\bf S}(\alpha, \beta_l, \gamma_l, \delta_l) (l=1, \ldots, L)$ is given by
        \begin{equation}
            \sum_l^L \epsilon_l
            \sim
            {\bf S}\left(\alpha, \frac{\sum_l^L\beta_l\gamma_l^\alpha}{\sum_l^L\gamma_l^\alpha}, \left(\sum_l^L\gamma_l^\alpha\right)^{\frac{1}{\alpha}}, \sum_l^L\delta_l\right)
            .
            \label{eq:sum_eps}
        \end{equation}
        For $L=2$, \eqref{eq:sum_eps} is directly derived from \Cref{prop:stabledist_transformation1}.
        Assume that \eqref{eq:sum_eps} holds for $\sum_l^{L-1}\epsilon_l$.
        Then, considering \Cref{prop:stabledist_transformation1}, the scale parameter of $\sum_l^{L-1} \epsilon_l + \epsilon_L$ is given by $(\sum_l^{L-1}\gamma_l^\alpha + \gamma_L^\alpha)^{\frac{1}{\alpha}} = (\sum_l^L\gamma_l^\alpha)^{\frac{1}{\alpha}}$.
        The skewness parameter is given by
        \begin{subequations}
        \begin{align}
            \frac{
            \frac{\sum_l^{L-1}\beta_l\gamma_l^\alpha}{\sum_l^{L-1}\gamma_l^\alpha}
            \cdot\sum_l^{L-1}\gamma_l^\alpha
            + \frac{\beta_L\gamma_L^\alpha}{\gamma_L^\alpha}
            \cdot\gamma_L^\alpha
            }{\sum_l^L\gamma_l^\alpha}
            &=
            \frac{\sum_l^{L-1}\beta_l\gamma_l^\alpha + \beta_L\gamma_L^\alpha}{\sum_l^L\gamma_l^\alpha}
            \\
            &= \frac{\sum_l^L\beta_l\gamma_l^\alpha}{\sum_l^L\gamma_l^\alpha}
            .
            \end{align}
        \end{subequations}
        The location parameter is given by $\sum_l^{L-1}\delta_l + \delta_L = \sum_l^{L}\delta_l$.
        This completes the proof of \eqref{eq:sum_eps}.
        
        Considering Taylor's theorem, because $f$ is linear over ${\bm \epsilon}$, we have 
        '
        \begin{multline}
            f(x_1; {\bm \epsilon}_1) - f(x_2; {\bm \epsilon}_2)
            = 
            f(x_1; \Delta) + \nabla_{\bm \epsilon} f(x_1)^\mathrm{T}({\bm \epsilon}_{1, k} - \Delta) 
            \\
            - f(x_2; \Delta)
            - \nabla_{\bm \epsilon} f(x_2)^\mathrm{T}({\bm \epsilon}_{2, k} - \Delta) 
            .
            \label{eq:taylor_diff}
        \end{multline}
        Consider the distribution of $\nabla_{\bm \epsilon} f(x)^\mathrm{T}({\bm \epsilon}_{1, k} - \Delta)$.
        From \Cref{prop:stabledist_transformation1}, 
        \begin{multline}
            g_m(x) \cdot (\epsilon_m - \delta_m)
            \sim
            {\bf S}\Big(\alpha, \mathrm{sign}(g_m(x))\cdot\beta_m, |g_m(x)|\cdot\gamma_m,
            \\
            -\frac{2}{\pi}\beta_m\gamma_m g_m(x)\log|g_m(x)|\cdot\ind{\alpha = 1}\Big)
            .
        \end{multline}
        Hence, using \eqref{eq:sum_eps},
        \begin{multline}
            \nabla_{\bm \epsilon} f(x)^\mathrm{T}({\bm \epsilon} - \Delta)
            =\sum_m^M g_m(x)\cdot(\epsilon_m - \delta_m) 
            \\
            \sim {\bf S}\left(\alpha, \beta'(x), \gamma'(x), \delta(x)'\cdot\ind{\alpha = 1}\right)
            \enspace.
        \end{multline}
        Moreover, according to \Cref{prop:stabledist_transformation1},
        \begin{multline}
            \nabla_{\bm \epsilon} f(x_1)^\mathrm{T}({\bm \epsilon}_1 - \Delta)
            -
            \nabla_{\bm \epsilon} f(x_2)^\mathrm{T}({\bm \epsilon}_2 - \Delta)
            \\
            \sim {\bf S}\left(\alpha, 
            \beta''(x_1, x_2)
            , \gamma''(x_1, x_2)
            , \delta''(x_1, x_2)\cdot\ind{\alpha = 1}\right)
            \enspace.
        \end{multline}
        Then, by adding $f(x_1; \Delta) - f(x_2; \Delta)$, we obtain
        \begin{multline}
            f(x_1; {\bm \epsilon}_1) - f(x_2; {\bm \epsilon}_2)
            \sim
            {\bf S}\Big(\alpha, 
            \beta''(x_1, x_2)
            , \gamma''(x_1, x_2)
            \\
            , f(x_1; \Delta) - f(x_2; \Delta) + \delta''(x_1, x_2)\cdot\ind{\alpha=1}\Big)
            .
        \end{multline}
        Again, \Cref{prop:stabledist_transformation1} and \eqref{eq:sum_eps} suggest that
        \begin{multline}
            \frac{1}{K}\sum_k^K 
            \rbra*{f(x_1; {\bm \epsilon}_1) - 
            f(x_2; {\bm \epsilon}_2)
            }
            \\
            \sim
            {\bf S}\Big(\alpha, 
            \beta''(x_1, x_2)
            , K^{\frac{1}{\alpha}-1}\cdot\gamma''(x_1, x_2), 
            \\
            f(x_1; \Delta) - f(x_2; \Delta) + \delta''(x_1, x_2)\cdot\ind{\alpha=1}\Big)
            .
        \end{multline}
        This completes the proof.
    \end{proof}

    \subsection{Proof of \Cref{theo:oep_stabledist}}\label{subsec:proof:theo:oep_stabledist}
    
    \begin{proof}
        From \Cref{lemma:diff_f} and \Cref{prop:stabledist_transformation1}, we have 
        \begin{multline}
            \frac{1}{K^{\frac{1}{\alpha}-1}\cdot\gamma''(x_1, x_2)}
            \cdot
            \left(
            \frac{1}{K}\sum_k^K  f(x_1; {\bm \epsilon}_{1, k}) - \frac{1}{K}\sum_k^Kf(x_2; {\bm \epsilon}_{2, k}) 
            \right)
            \\
            \sim
            {\bm S}\Bigg(
            \alpha,\beta''(x_1, x_2), 
            1,
            \frac{f(x_1; \Delta) - f(x_2; \Delta)}{K^{\frac{1}{\alpha}-1}\cdot\gamma''(x_1, x_2)}
           \\ 
             + \left(
             \frac{\delta''(x_1, x_2)}{\gamma''(x_1, x_2)}
             + \frac{2}{\pi}\beta''(x_1, x_2)\cdot\log(\gamma''(x_1, x_2))
             \right)\cdot\ind{\alpha = 1}
            \Bigg)
            .
        \end{multline}
        $K^{\frac{1}{\alpha}-1}$ disappears in the second item of the location parameter for $\alpha=1$ and $\gamma''(x_1, x_2)>0$ for any $x_1, x_2$.
        As $\epsilon^\mathrm{AVE} + \frac{f(x_1; \Delta) - f(x_2; \Delta)}{K^{\frac{1}{\alpha}-1}\cdot\gamma''(x_1, x_2)}$ and $\frac{1}{K^{\frac{1}{\alpha}-1}\cdot\gamma''(x_1, x_2)}
            \cdot
            \left(
            \frac{1}{K}\sum_k^K  f(x_1; {\bm \epsilon}_{1, k}) - \frac{1}{K}\sum_k^Kf(x_2; {\bm \epsilon}_{2, k}) 
            \right)$ are equivalent in distribution, we have 
            \begin{subequations}
            \begin{align}
                \MoveEqLeft[0]\Pr_{\bm \epsilon}\left[\widehat\eta_\epsilon^\mathrm{AVE}(x_1, x_2) = \eta_\Delta(x_1, x_2)\right]
                \\
                &=\begin{aligned}[t]
                \Pr_{\bm \epsilon}\Bigg[
                    &\left( 
                    \frac{1}{K}\sum_k^K \rbra*{f(x_1, {\bm \epsilon}_{1, k}) - f(x_2, {\bm \epsilon}_{2, k})}
                    \right)
                    \\
                    &\cdot
                    (f(x_1; \Delta) - f(x_2; \Delta)) > 0
                \Bigg]
                \end{aligned}
                \\
                &\begin{multlined}
                =\Pr_{\bm \epsilon}\left[
                    \epsilon^\mathrm{AVE} > -\frac{f(x_1; \Delta) - f(x_2; \Delta)}{K^{\frac{1}{\alpha}-1}\cdot\gamma''(x_1, x_2)}
                \right]
                \cdot\ind{\eta_\Delta(x_1, x_2)>0}
                \\
                + 
                \Pr_{\bm \epsilon}\left[
                    \epsilon^\mathrm{AVE} < -\frac{f(x_1; \Delta) - f(x_2; \Delta)}{K^{\frac{1}{\alpha}-1}\cdot\gamma''(x_1, x_2)}
                \right]
                \cdot\ind{\eta_\Delta(x_1, x_2)<0}
                \end{multlined}
                \\
                &=\begin{aligned}[t]
                &\Pr_{\bm \epsilon}\left[
                    -\eta_\Delta(x_1, x_2)\cdot
                    \epsilon^\mathrm{AVE} < \frac{\abs{f(x_1; \Delta) - f(x_2; \Delta)}}{K^{\frac{1}{\alpha}-1}\cdot\gamma''(x_1, x_2)}
                \right]
                \\
                &\cdot\ind{\eta_\Delta(x_1, x_2)>0}
                \\
                &+ 
                \Pr_{\bm \epsilon}\left[
                    -\eta_\Delta(x_1, x_2)\cdot
                    \epsilon^\mathrm{AVE} < \frac{\abs{f(x_1; \Delta) - f(x_2; \Delta)}}{K^{\frac{1}{\alpha}-1}\cdot\gamma''(x_1, x_2)}
                \right]
                \\
                &\cdot\ind{\eta_\Delta(x_1, x_2)<0}
                \end{aligned}
                \\
                &=\begin{aligned}[t]
                &\Pr_{\bm \epsilon}\left[
                    -\eta_\Delta(x_1, x_2)\cdot
                    \epsilon^\mathrm{AVE} < \frac{\abs{f(x_1; \Delta) - f(x_2; \Delta)}}{K^{\frac{1}{\alpha}-1}\cdot\gamma''(x_1, x_2)}
                \right]
                \\
                &\cdot\ind{\eta_\Delta(x_1, x_2)\neq 0}
                .
                \end{aligned}
                \end{align}
                \end{subequations}
            This completes the proof.
        \end{proof}

    \subsection{Proof of \Cref{cor:oep_stabledist_sym}}\label{subsec:proof:cor:oep_stabledist_sym}
    \begin{proof}
        Let $\beta_m=0, \forall m$, $\beta''(x_1, x_2)=0$, and $\delta''(x_1, x_2)=0$ for any $x_1, x_2$.
        We have 
        \begin{equation}
            -\eta_\Delta(x_1, x_2)\cdot \epsilon^\mathrm{AVE}
            \sim
            {\bf S}(\alpha, 0, 1, 0)
            .
        \end{equation}
        This completes the proof.
    \end{proof}

    \subsection{Proof of \Cref{prop:symmetric}}\label{apdx:prop:symmetric}
    \begin{proof}
        Let $X_1 = f(x_1; {\bm \epsilon}_1) - m_f(x_1)$ and $X_2 = f(x_2; {\bm \epsilon}_2) - m_f(x_2)$. 
        They are independent, and their distributions are symmetric around the origin.
        Let $Z = \epsilon^f(x_1, x_2) - (m_f(x_1) - m_f(x_2))$. 
        Hence, we have $Z = X_1 - X_2$. 
        If $Z$ is symmetric around the origin, $\epsilon^f(x_1, x_2)$ is symmetric around $m_f(x_1) - m_f(x_2)$, and the center of the median of $\epsilon^f(x_1, x_2)$ is $m_f(x_1) - m_f(x_2)$. Therefore, it is sufficient to demonstrate that $Z$ is symmetric around the origin.
        On the other hand, it is well-known that a sum of independent random variables that are both symmetric around the origin is symmetric around the origin~\cite{grimmett2020probability}.
        Therefore, the center of the median of $\epsilon^f(x_1, x_2)$ is $m_f(x_1) - m_f(x_2)$.
        This completes the proof.
    \end{proof}

\new{
\subsection{Proof of \Cref{prop:identity}}\label{apdx:prop:identity}
\begin{proof}
Let $\xi_x = f(x; \bm{\epsilon}_x) - m_f(x)$ and $\xi_y = f(y; \bm{\epsilon}_y) - m_f(y)$, where $\bm{\epsilon}_x$ and $\bm{\epsilon}_y$  are i.i.d.
Then, $\xi_x$ and $\xi_y$ are i.i.d.\ as well. 
Because they are i.i.d., $\xi_x - \xi_y$ is symmetric around $0$. Therefore, the center of the median of $\xi_x - \xi_y$ is $0$. 
Now, consider the center of the median of $f(x; \bm{\epsilon}_x) - f(y; \bm{\epsilon}_y)$, namely $m_f(x, y)$. 
Because $f(x; \bm{\epsilon}_x) - f(y; \bm{\epsilon}_y) = m_f(x) - m_f(y) + (\xi_x - \xi_y)$, $m_f(x, y)$ is the sum of $m_f(x) - m_f(y)$ and the center of the median of $\xi_x - \xi_y$, the latter of which is $0$. Hence, we have $m_f(x, y) = m_f(x) - m_f(y)$. 
\end{proof}
}

\new{

\subsection{Proof of \Cref{prop:uniqueness}}\label{apdx:prop:uniqueness}
\begin{proof}
Let $X = f(x; \bm{\epsilon}) - m_f(x)$.
Then, $X$ is a random variable and the center of its median is $0$. 
Because $\xi$ has a positive probability density over $\R$, it has a unique median $m_\xi$. 
Let $Y = \xi - m_\xi$. Obviously, its unique median is $0$. 
Because $X$ and $Y$ are independent and $Y$ is absolutely continuous, $Z = X + Y$ is absolutely continuous.
Let $F_x$ be the cumulative distribution function of $X$ and $p_y$ be the probability density function of $Y$.
Let the probability density function of $Z$ be denoted by $p_z$. 
Let $(L, U]$ be an interval with $F_x(U) - F_x(L) > 0$. 
Then, for any $z \in \R$, we have $\min_{t \in [z - U, z - L]} p_y(t) > 0$ as $p_y$ is continuous and strictly positive, and
\begin{subequations}
\begin{align}
p_z(\xi) 
&= \int_{x \in \R} p_y(z - x) \mathrm{d}F_x(x)
\\
&\geq \int_{x \in (L, U]} p_y(z - x) \mathrm{d}F_x(x)
\\
&\geq \min_{t \in [z - U, z - L]} p_y(t) (F_x(U) - F_x(L))
\\
& > 0 \enspace.
\end{align}
\end{subequations}
Therefore, $p_z(z) > 0$ for all $z \in \R$. 
Because $f(x; \epsilon) = Z + m_f(x) + m_\xi$, its probability density is $p(t) = p_z(t - m_f(x) + m_\xi)$. 
Then, it is easy to see that $p(t) > 0$ for all $t \in \R$. 
Because $p$ is the derivative of the cumulative distribution function of $f(x; \epsilon)$, the cumulative distribution function is differentiable. Therefore, \Cref{asm:uniqueness} is satisfied.
\end{proof}
}

        \subsection{Proof of \Cref{prop:SA_symmetric}}\label{subsec:proof:prop:SA_symmetric}
        \begin{proof}
            We first show that \eqref{eq:signproduct} under \Cref{asm:symmetric}.
            Let $Z = \epsilon^f(x_1, x_2) - m_f(x_1, x_2)$. 
        We have
        \begin{subequations}
            \begin{align}
            \MoveEqLeft[2]\E\left[\mathrm{sign}(\epsilon^f(x_1, x_2))\right]\notag
            \\
            &=
            \Pr\sbra*{\epsilon^f(x_1, x_2)>0}
            - \Pr\sbra*{\epsilon^f(x_1, x_2)<0}
            \\
            &=
            \Pr\sbra*{Z > - m_f(x_1, x_2)}
            - \Pr\sbra*{Z < - m_f(x_1, x_2)}.\label{eq:signproductproof}
            \end{align}
        \end{subequations}
        Clearly, the first and second terms are no smaller and greater than $\frac{1}{2}$, respectively, if $m_f(x_1, x_2) > 0$. Under \Cref{asm:symmetric}, we have $m_f(x_1, x_2) > 0 \Leftrightarrow m_f(x_1) > m_f(x_2)$. Therefore, $\E\left[\mathrm{sign}(\epsilon^f(x_1, x_2))\right] \geq 0$ if $m_f(x_1) > m_f(x_2)$.
        Similarly, if $m_f(x_1) < m_f(x_2)$, $\E\left[\mathrm{sign}(\epsilon^f(x_1, x_2))\right] \leq 0$.
        Thus, it is straightforward to obtain \eqref{eq:signproduct}.

        Next, we prove \eqref{eq:signequality}. 
        Let $X_1 = f(x_1; {\bm \epsilon}_1) - m_f(x_1)$ and $X_2 = f(x_2; {\bm \epsilon}_2) - m_f(x_2)$. 
        Then, we have $Z = X_1 - X_2$. 
        Let $\phi_1$, $\phi_2$, and $\phi_z$ be the distribution functions of $X_1$, $X_2$, and $Z$, respectively. 
        Suppose that $\phi_z(z) > \phi_z(0) = \frac{1}{2}$ and $\phi_z(-z) < \phi_z(0) = \frac{1}{2}$ for any $z > 0$.
        Then, if $m_f(x_1, x_2) > 0$, we have
        \begin{equation}
            \Pr\sbra*{Z > - m_f(x_1, x_2)} 
            = 1 - \phi_z(- m_f(x_1, x_2)) > \frac{1}{2}
        \end{equation}
        and 
        \begin{equation}
            - \Pr\sbra*{Z < - m_f(x_1, x_2)} 
            \leq \phi_z(- m_f(x_1, x_2)) \leq \frac{1}{2}. 
        \end{equation}
        Thus, 
            $\E\left[\mathrm{sign}(\epsilon^f(x_1, x_2))\right] > 0$.
        Analogously, 
        if $m_f(x_1, x_2) < 0$, 
            $\E\left[\mathrm{sign}(\epsilon^f(x_1, x_2))\right] < 0$.
        Hence, it suffices to show that $\phi_z(z) > \phi_z(0) = \frac{1}{2}$ and $\phi_z(-z) < \phi_z(0) = \frac{1}{2}$ for any $z > 0$ under \Cref{asm:symmetric} and \Cref{asm:uniqueness}. 
We already know that $Z$ is symmetric around the origin and that the median is located at the origin. 
Therefore, we have $\phi_z(z) \geq \frac{1}{2}$ for any $z \geq 0$, $\phi(z) \leq \frac{1}{2}$ for $z < 0$, and $\phi_z(z) = 1 - \lim_{\rho \downarrow 0}\phi_z(-z - \rho)$. 
In other words, if we show $\phi(z) < \frac{1}{2}$ for any $z < 0$, then $\phi(z) = 1 - \lim_{\rho \downarrow 0}\phi_z(-z - \rho) \geq 1 - \phi(-z) > \frac{1}{2}$ for any $z > 0$. 

We prove that $\phi(z) < \frac{1}{2}$ for any $z < 0$ holds. 
Under \Cref{asm:uniqueness}, there exists $\delta > 0$ such that $\phi_1$ and $\phi_2$ are continuous and strictly increasing in interval $[-\delta, \delta]$. 
Let $- \delta \leq z < - \rho < 0$. We have
\begin{subequations}
\begin{align}
            \MoveEqLeft[0]\phi_z(z)\notag\\
            &= \int^\infty_{-\infty} \phi_1(z - t) \mathrm{d}\phi_2(t)
\\
            &= \begin{aligned}[t]
            &\int^\infty_{-\infty} \phi_1(-\rho-t) \mathrm{d}\phi_2(t) 
            \\
            &+ \int^\infty_{-\infty} \left( \phi_1(z - t) - \phi_1(-\rho-t) \right)  \mathrm{d}\phi_2(t)
            \end{aligned}
\\
            &= \phi_z(-\rho) + \int^\infty_{-\infty} \left( \phi_1(z - t) - \phi_1(-\rho-t) \right) \mathrm{d}\phi_2(t)
            \\
            &\leq \phi_z(-\rho) + \int_{z}^{-\rho} \left( \phi_1(z - t) - \phi_1(-\rho-t) \right)   \mathrm{d}\phi_2(t)
            \\
            &\leq \phi_z(-\rho) + \int_{z}^{-\rho} \left( \max_{z \leq t \leq -\rho}\phi_{1}(z - t) - \phi_1(-\rho-t) \right) \mathrm{d}\phi_2(t)
            \\
            &=\begin{aligned}[t]
            &\phi_z(-\rho) 
            \\
            &+  \left( \max_{z \leq t \leq -\rho}\phi_{1}(z - t) - \phi_1(-\rho-t) \right) \left( \phi_2(-\rho) - \phi_2(z)\right)
    .
            \end{aligned}
            \end{align}   
\end{subequations}
Here, we have the Lebesgue--Stieltjes integral.
For the first inequality, we consider that $ \phi_1(z - t) - \phi_1(-\rho-t) \leq 0$ for all $t \in \R$ because $\phi_1$ increases at least weakly.
For the second inequality, we consider that $ \phi_1(z - t) - \phi_1(-\rho-t) < 0$ for all $t \in [z, -\rho]$ because $-\delta < z + \rho \leq  z - t < -\rho - t \leq - \rho - z < \delta$ and $\phi_1$ is strictly increasing in $[-\delta, \delta]$.
We also consider that $\phi_{1}(z - t) - \phi_1(-\rho-t)$ is continuous on $t \ in [z, -\rho]$ and its maximum exists in $[z, -\rho]$. 

The desired inequality is derived by evaluating each component on the rightmost side of the inequality. 
For the first term, $\phi_z(- \rho) \leq \frac{1}{2}$. 
For the component of $\phi_2$, we have $\phi_2(-\rho) - \phi_2(z) > 0$ because $\phi_2$ is strictly increasing in $[-\delta, \delta]$. 
The strict negativity of component $\phi_1$ is derived as follows.
For $z < t < -\rho$, we have $\phi_{1}(z - t) < \phi_1(0) = \frac{1}{2}$ and $\phi_1(-\rho-t) > \phi_1(0) = \frac{1}{2}$. 
For $t=z$, $\phi_1(z-t) < \phi_1(0) = \frac{1}{2}$ and $\phi_1(-\rho-t) = \phi_1(-\rho-z) > \phi_1(0) = \frac{1}{2}$.
For $t=-\rho$, $\phi_1(z-t) = \phi_1(z + \rho) < \frac{1}{2}$ and $\phi_1(-\rho-t) = \phi_1(0) = \frac{1}{2}$.
Thus, $\phi_{1}(z - t) - \phi_1(-\rho-t) < 0$ for all $t \in [z, -\rho]$. 
Altogether, we have $\phi_z(z) < \frac{1}{2}$ for $z \in [-\delta, 0)$, which immediately leads to $\phi_z(z) < \frac{1}{2}$ for all $z < 0$ from the monotonicity of the cumulative distribution function. 
\end{proof}   

    \subsection{Proof of \Cref{theo:oep_stabledist_sign}}\label{subsec:proof:theo:oep_stabledist_sign}
    \begin{proof}
        Let $(x_1, x_2)$ be a pair of points in $\R^D$ that satisfy $\eta_\mathrm{med}(x_1, x_2) > 0$.
        In addition, let $\epsilon^f_k(x_1, x_2)$, where $k=1, \ldots, K$, be i.i.d.\ copies of $\epsilon^f(x_1, x_2)$.
        We have 
        \begin{subequations}
        \begin{align}
            \MoveEqLeft[0]\Pr_{\bm \epsilon}\left[\widehat\eta_{\bm \epsilon}^\mathrm{SA}(x_1, x_2)
            = \eta_\mathrm{med}(x_1, x_2)\right]\notag
            \\
            &=\Pr_{\bm \epsilon}\left[
                \left(
                \frac{1}{K}\sum_k^K
                \mathrm{sign}(f(x_1;{\bm \epsilon}_{1, k}) - f(x_2; {\bm \epsilon}_{2, k}))
                \right)
                \eta_\mathrm{med}(x_1, x_2)
                >0
            \right]
            \\
            &=
            \Pr_{\bm \epsilon}\left[
                \frac{1}{K}\sum_k^K
                \mathrm{sign}(\epsilon^f_k(x_1, x_2))
                >0
            \right]
            \\
            &=
            1- \Pr_{\bm \epsilon}\left[
                \frac{1}{K}\sum_k^K
                \mathrm{sign}(\epsilon^f_k(x_1, x_2))
                \leq 0
            \right]           
            .
        \end{align}
        \end{subequations}
        Next, we evaluate the upper bound of $\Pr_{\bm \epsilon}\left[
                \frac{1}{K}\sum_k^K
                \mathrm{sign}(\epsilon^f_k(x_1, x_2))
                \leq 0\right]$.
        As $\left|\mathrm{sign}\left(\epsilon^f(x_1, x_2)\right)\right|\leq1$ surely,
        both $\E\left[\mathrm{sign}\left(\epsilon^f(x_1, x_2)\right)\right]$ and
        $\Var\left[\mathrm{sign}\left(\epsilon^f(x_1, x_2)\right)\right]$ are finite.
        Under \Cref{asm:symmetric} and condition $\eta_\mathrm{med}(x_1, x_2) > 0$ and considering \Cref{prop:SA_symmetric}, we have $\E[\mathrm{sign}\left(\epsilon^f(x_1, x_2)\right)] \geq 0$. 
        Hence,
        \begin{subequations}
        \begin{align}
            \MoveEqLeft[0]\Pr_{\bm \epsilon}\left[
                \frac{1}{K}\sum_k^K
                \mathrm{sign}\left(\epsilon^f_k(x_1, x_2)\right)
                \leq0
            \right]         \notag
            \\
            &=
            \Pr_{\bm \epsilon}\Bigg[
                K\cdot\E\left[\mathrm{sign}\left(\epsilon^f(x_1, x_2)\right)\right]
                -
                \sum_k^K
                \mathrm{sign}\left(\epsilon^f_k(x_1, x_2)\right)
                \\
                &
                \enspace
                \enspace
                \geq
                K\cdot\E[\mathrm{sign}\left(\epsilon^f(x_1, x_2)\right)]\Bigg]
            .
        \end{align}
        \end{subequations}
        Considering Hoeffding's inequality, the righthand side of the above inequality is bounded from above by
        \begin{equation}
            \exp\rbra*{-\frac{K\cdot\rbra*{\E\left[\mathrm{sign}\left(\epsilon^f(x_1, x_2)\right)\right]}^2}{2}}
            .
        \end{equation}
        Therefore, for $\eta(x_1, x_2)>0$, we have 
        \begin{multline}
            \Pr_{\bm \epsilon}\left[\widehat\eta_{\bm \epsilon}^\mathrm{SA}(x_1, x_2)
            = \eta_\mathrm{med}(x_1, x_2)\right]
            \\
            \geq
            1-
             \exp\rbra*{-\frac{K\cdot\rbra*{\E\left[\mathrm{sign}\left(\epsilon^f(x_1, x_2)\right)\right]}^2}{2}}           
            .
        \end{multline}
        Similarly, assuming that $\eta(x_1, x_2) <0$, we obtain the same inequality to that obtained for $\eta(x_1, x_2)>0$.
        This completes the proof.
        \end{proof}
    

\begin{thebibliography}{10}

\bibitem{jastrebski2006improving}
Improving evolution strategies through active covariance matrix adaptation.
\newblock In {\em 2006 IEEE international conference on evolutionary
  computation}, pages 2814--2821. IEEE, 2006.

\bibitem{akimoto2020diagonal}
Diagonal acceleration for covariance matrix adaptation evolution strategies.
\newblock {\em Evolutionary computation}, 28(3):405--435, 2020.

\bibitem{agresti2010analysis}
Alan Agresti.
\newblock {\em Analysis of ordinal categorical data}, volume 656.
\newblock John Wiley \& Sons, 2010.

\bibitem{ahrari2022revisiting}
Ali Ahrari, Saber Elsayed, Ruhul Sarker, Daryl Essam, and Carlos A~Coello
  Coello.
\newblock Revisiting implicit and explicit averaging for noisy optimization.
\newblock {\em IEEE Transactions on Evolutionary Computation}, 2022.

\bibitem{akimoto2022ode}
Youhei Akimoto, Anne Auger, and Nikolaus Hansen.
\newblock An ode method to prove the geometric convergence of adaptive
  stochastic algorithms.
\newblock {\em Stochastic Processes and their Applications}, 145:269--307,
  2022.

\bibitem{arnold2006general}
Dirk~V Arnold and H-G Beyer.
\newblock A general noise model and its effects on evolution strategy
  performance.
\newblock {\em IEEE Transactions on Evolutionary Computation}, 10(4):380--391,
  2006.

\bibitem{arnold2001local}
Dirk~V Arnold and Hans-Georg Beyer.
\newblock Local performance of the ($\mu$/$\mu$i, $\lambda$)-es in a noisy
  environment.
\newblock In {\em Foundations of Genetic Algorithms 6}, pages 127--141.
  Elsevier, 2001.

\bibitem{auger2016linear}
Anne Auger and Nikolaus Hansen.
\newblock Linear convergence of comparison-based step-size adaptive randomized
  search via stability of markov chains.
\newblock {\em SIAM Journal on Optimization}, 26(3):1589--1624, 2016.

\bibitem{beyer2006functions}
H-G Beyer and Bernhard Sendhoff.
\newblock Functions with noise-induced multimodality: a test for evolutionary
  robust optimization-properties and performance analysis.
\newblock {\em IEEE Transactions on Evolutionary Computation}, 10(5):507--526,
  2006.

\bibitem{beyer1993toward}
Hans-Georg Beyer.
\newblock Toward a theory of evolution strategies: Some asymptotical results
  from the (1,+ $\lambda$)-theory.
\newblock {\em Evolutionary computation}, 1(2):165--188, 1993.

\bibitem{beyer2000evolutionary}
Hans-Georg Beyer.
\newblock Evolutionary algorithms in noisy environments: Theoretical issues and
  guidelines for practice.
\newblock {\em Computer methods in applied mechanics and engineering},
  186(2-4):239--267, 2000.

\bibitem{beyer2007robust}
Hans-Georg Beyer and Bernhard Sendhoff.
\newblock Robust optimization--a comprehensive survey.
\newblock {\em Computer methods in applied mechanics and engineering},
  196(33-34):3190--3218, 2007.

\bibitem{casella2021statistical}
George Casella and Roger~L Berger.
\newblock {\em Statistical inference}.
\newblock Cengage Learning, 2021.

\bibitem{doerr2019resampling}
Benjamin Doerr and Andrew~M Sutton.
\newblock When resampling to cope with noise, use median, not mean.
\newblock In {\em Proceedings of the Genetic and Evolutionary Computation
  Conference}, pages 242--248, 2019.

\bibitem{dong2019efficient}
Yinpeng Dong, Hang Su, Baoyuan Wu, Zhifeng Li, Wei Liu, Tong Zhang, and Jun
  Zhu.
\newblock Efficient decision-based black-box adversarial attacks on face
  recognition.
\newblock In {\em Proceedings of the IEEE/CVF Conference on Computer Vision and
  Pattern Recognition}, pages 7714--7722, 2019.

\bibitem{fujii2018cma}
Garuda Fujii, Masayuki Takahashi, and Youhei Akimoto.
\newblock Cma-es-based structural topology optimization using a level set
  boundary expression―application to optical and carpet cloaks.
\newblock {\em Computer Methods in Applied Mechanics and Engineering},
  332:624--643, 2018.

\bibitem{gabrel2014recent}
Virginie Gabrel, C{\'e}cile Murat, and Aur{\'e}lie Thiele.
\newblock Recent advances in robust optimization: An overview.
\newblock {\em European journal of operational research}, 235(3):471--483,
  2014.

\bibitem{ghadimi2013stochastic}
Saeed Ghadimi and Guanghui Lan.
\newblock Stochastic first-and zeroth-order methods for nonconvex stochastic
  programming.
\newblock {\em SIAM Journal on Optimization}, 23(4):2341--2368, 2013.

\bibitem{grimmett2020probability}
Geoffrey Grimmett and David Stirzaker.
\newblock {\em Probability and random processes}.
\newblock Oxford university press, 2020.

\bibitem{hansen2006cma}
Nikolaus Hansen.
\newblock The cma evolution strategy: a comparing review.
\newblock {\em Towards a new evolutionary computation: Advances in the
  estimation of distribution algorithms}, pages 75--102, 2006.

\bibitem{hansen2003reducing}
Nikolaus Hansen, Sibylle~D M{\"u}ller, and Petros Koumoutsakos.
\newblock Reducing the time complexity of the derandomized evolution strategy
  with covariance matrix adaptation (cma-es).
\newblock {\em Evolutionary computation}, 11(1):1--18, 2003.

\bibitem{hansen2008method}
Nikolaus Hansen, Andr{\'e}~SP Niederberger, Lino Guzzella, and Petros
  Koumoutsakos.
\newblock A method for handling uncertainty in evolutionary optimization with
  an application to feedback control of combustion.
\newblock {\em IEEE Transactions on Evolutionary Computation}, 13(1):180--197,
  2008.

\bibitem{hansen2001completely}
Nikolaus Hansen and Andreas Ostermeier.
\newblock Completely derandomized self-adaptation in evolution strategies.
\newblock {\em Evolutionary computation}, 9(2):159--195, 2001.

\bibitem{jin2005evolutionary}
Yaochu Jin and J{\"u}rgen Branke.
\newblock Evolutionary optimization in uncertain environments-a survey.
\newblock {\em IEEE Transactions on evolutionary computation}, 9(3):303--317,
  2005.

\bibitem{kriest2017calibrating}
Iris Kriest, Volkmar Sauerland, Samar Khatiwala, Anand Srivastav, and Andreas
  Oschlies.
\newblock Calibrating a global three-dimensional biogeochemical ocean model
  (mops-1.0).
\newblock {\em Geoscientific Model Development}, 10(1):127--154, 2017.

\bibitem{nelder1965simplex}
John~A Nelder and Roger Mead.
\newblock A simplex method for function minimization.
\newblock {\em The computer journal}, 7(4):308--313, 1965.

\bibitem{nolan2020stable}
John~P Nolan.
\newblock {\em Univariate stable distributions}.
\newblock Springer, 2020.

\bibitem{qian2018effectiveness}
Chao Qian, Yang Yu, Ke~Tang, Yaochu Jin, Xin Yao, and Zhi-Hua Zhou.
\newblock On the effectiveness of sampling for evolutionary optimization in
  noisy environments.
\newblock {\em Evolutionary computation}, 26(2):237--267, 2018.

\bibitem{rakshit2017noisy}
Pratyusha Rakshit, Amit Konar, and Swagatam Das.
\newblock Noisy evolutionary optimization algorithms--a comprehensive survey.
\newblock {\em Swarm and Evolutionary Computation}, 33:18--45, 2017.

\end{thebibliography}

\section{Biography Section}
 
 \vspace{11pt}

%

\vspace{11pt}
\vfill

\end{document}